\PassOptionsToPackage{table}{xcolor}
\documentclass[sigconf, screen]{acmart}
\renewcommand\footnotetextcopyrightpermission[1]{}
\AtBeginDocument{
  }

\setcopyright{none}
\copyrightyear{2018}
\acmYear{2018}
\acmDOI{XXXXXXX.XXXXXXX}
\acmConference[Conference acronym 'XX]{Make sure to enter the correct
  conference title from your rights confirmation email}{June 03--05,
  2018}{Woodstock, NY}

\acmISBN{978-1-4503-XXXX-X/2018/06}

\usepackage{colortbl}
\usepackage{booktabs}
\usepackage{multirow}
\usepackage{tabularx}
\usepackage[skins]{tcolorbox}
\newcommand{\mnt}[1]{\texttt{max\_\allowbreak new\_\allowbreak tokens=#1}}
\definecolor{groupgray}{gray}{0.92}
\definecolor{hlblue}{RGB}{218,232,252}
\definecolor{hlblue_light}{RGB}{240,246,255}
\definecolor{blockbg}{gray}{0.98}
\begin{document}
\title{Geo3R: Mitigating Spatial Reasoning Hallucination in Multimodal Large Language Models}

\author{Mingyu Wang}
\affiliation{
  \institution{Peking University}
  \city{Beijing}
  \country{China}
}

\author{Weilin Jin}
\affiliation{
  \institution{Peking University}
  \city{Beijing}
  \country{China}
}

\author{Wenbo Li}
\authornote{Project leader.}
\affiliation{
  \institution{Joy Future Academy}
  \city{Beijing}
  \country{China}
}

\author{Haoyang Huang}
\affiliation{
  \institution{Joy Future Academy}
  \city{Beijing}
  \country{China}
}

\author{Tong Jia}
\authornote{Corresponding author.}
\affiliation{
  \institution{Peking University}
  \city{Beijing}
  \country{China}
}

\author{Ying Li}
\authornotemark[2]
\affiliation{
  \institution{Peking University}
  \city{Beijing}
  \country{China}
}

\renewcommand{\shortauthors}{Wang et al.}

\begin{abstract}
Despite remarkable progress in visual understanding, Multimodal Large Language Models (MLLMs) remain prone to hallucinations when reasoning about spatial relationships, often producing judgments that contradict the true 3D structure of the scene.
Though several existing works have proposed to mitigate hallucinations, our analysis indicates that they show limited effectiveness in spatial reasoning, as they fail to bridge the fundamental gap between 2D visual representations and 3D spatial reality.
Based on this finding, we define hallucinations arising from insufficient spatial structure modeling as spatial reasoning hallucination, a subcategory of relation hallucination that existing mitigation methods fail to address.
We further identify three typical scenarios where such hallucinations frequently occur: perspective effects, object orientation, and viewpoint changes.
To this end, we propose Geo3R, a training-free, plug-and-play framework that incorporates geometric evidence and structured 3D reasoning to mitigate spatial reasoning hallucination.
Experiments on three benchmarks, covering 18 tasks across all three scenarios, show that Geo3R substantially reduces spatial reasoning hallucination across diverse MLLMs without additional training, outperforming existing models and methods.
\end{abstract}




\received{20 February 2007}
\received[revised]{12 March 2009}
\received[accepted]{5 June 2009}

\maketitle

\section{Introduction}
\label{sec:introduction}

\begin{figure}[t]
\centering
\includegraphics[width=0.9\columnwidth]{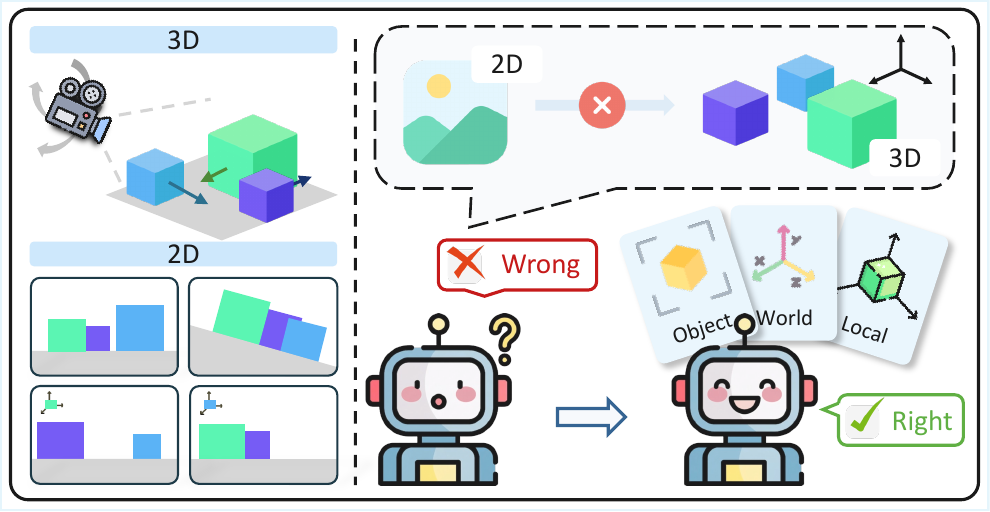}
\caption{The gap between 2D representations and 3D reality makes spatial reasoning tasks challenging for MLLMs.}
\Description{A teaser figure showing six spatial reasoning challenges from camera geometry (left), the gap between 2D representations and 3D reality (right-top), and how Geo3R uses structured geometric cards to enable correct spatial reasoning (right-bottom).}
\label{fig:intro}
\end{figure}

Multimodal Large Language Models (MLLMs) have achieved remarkable progress in visual understanding, multimodal question answering, and complex reasoning tasks~\cite{llava-1.5,yang2025qwen3,team2023gemini,singh2025openai}.
However, these models continue to suffer from hallucination, generating content that contradicts the visual input~\cite{bai2024hallucination}.
Prior work typically categorizes hallucinations into three types: existence, attribute, and relation~\cite{tu2025ode,trihe}.
Among these, relation hallucination is particularly challenging, as it requires understanding relationships between objects~\cite{zheng2025reefknot, trihe}.
Such hallucinations can significantly compromise model reliability in real-world applications such as embodied navigation, robotic manipulation, and autonomous driving.

Within relation hallucination, spatial relations introduce an additional challenge by requiring models to infer 3D spatial relations from 2D images.
Prior studies have shown that this challenge leads to severe spatial relation hallucinations in MLLMs, with high hallucination rates in depth ordering, object orientation, and relative distance tasks, while accuracy on viewpoint-dependent judgments remains close to chance level~\cite{li2025viewspatial,ma20253dsrbench}.
As illustrated in Figure~\ref{fig:intro}, the fundamental cause is the discrepancy between 2D visual representations and 3D spatial reality.
For instance, perspective projection can cause a physically higher but more distant object to appear lower in the image than a nearer one, camera tilt distorts the apparent height and position of objects, and viewpoint changes alter the perceived spatial arrangement between objects.
This has motivated the development of models specifically designed for spatial understanding~\cite{cheng2024spatialrgpt,batra2025spatialthinker,Spatial-ssrl}. 
However, these models rely on additional training with specialized spatial data, which incurs substantial cost, limits scalability, and may yield inconsistent improvements across different forms of spatial reasoning.

Compared to these specialized models, training-free hallucination mitigation methods~\cite{huang2024opera,trihe,zheng2025reefknot,chen2025spatial} enable lightweight deployment and easy integration with existing MLLMs.
However, it remains unclear whether such methods are effective for spatial reasoning. 
To examine this, we evaluate these methods on spatial reasoning tasks (Section~\ref{sec:preliminary}).
The results show that these methods yield negligible or even negative improvements, as they are primarily designed to improve semantic alignment between visual content and generated text, typically by adjusting attention or decoding based on existing 2D visual representations.
As a result, while they may help reduce hallucinations caused by visual-semantic misalignment, they remain insufficient for addressing spatial hallucinations rooted in the gap between 2D visual representations and 3D spatial reality.

Motivated by this analysis, we define \textbf{spatial reasoning hallucination} as a subcategory of relation hallucination caused not by visual-textual misalignment, but by insufficient spatial structure modeling.
Specifically, we identify three representative scenarios where the gap between 2D and 3D is especially pronounced, namely perspective effects, object orientation, and viewpoint changes.
Across all three scenarios, the key challenge lies in enabling 3D reasoning from 2D visual input alone.

To address this challenge, we propose \textbf{Geo3R} (\textbf{Geo}metric \textbf{3}D \textbf{R}easoning), a training-free, plug-and-play framework that augments any MLLM with explicit geometric reasoning from a single image.
The key idea is to recover 3D geometric evidence from 2D visual input and present it as structured cues that guide the MLLM toward spatially grounded answers.
Geo3R comprises three sequential stages, \textit{Visual Grounding}, \textit{Geometric Evidence Extraction}, and \textit{Geometric-Augmented Spatial Reasoning}.
The first stage identifies task-relevant objects.
Based on the detected objects, the second stage leverages pretrained geometric estimation tools to construct multi-space 3D representations spanning camera, gravity-aligned world, and object-local coordinate frames.
The third stage then organizes the extracted evidence into structured geometric cards to guide the MLLM's reasoning.
Experiments on three benchmarks, covering 18 tasks across all three scenarios, show that Geo3R improves the average accuracy of Gemini-3-Flash by 7.06\% and Qwen3-VL-8B by 10.90\%, with both surpassing GPT-5 after augmentation.

Our contributions are as follows.
\begin{itemize}
\setlength{\itemsep}{1pt}\setlength{\parsep}{0pt}\setlength{\topsep}{0pt}
    \item We analyze the failure of existing training-free hallucination mitigation methods on spatial reasoning tasks and define spatial reasoning hallucination, identifying three representative scenarios, namely perspective effects, object orientation, and viewpoint changes.
    \item We propose Geo3R, a training-free, plug-and-play framework that bridges the gap between 2D visual representations and 3D spatial reality through multi-space geometric representations and structured geometric cards, enabling explicit spatial reasoning for any MLLM.
    \item Extensive experiments on three benchmarks demonstrate that Geo3R substantially improves diverse MLLMs, outperforming existing models and methods.
\end{itemize}

\section{Related Work}
\label{sec:related}

\subsection{Spatial Reasoning in MLLMs}
Despite the remarkable progress of MLLMs in visual understanding, spatial reasoning remains a significant challenge~\cite{li2025viewspatial,ma20253dsrbench,tong2024cambriancvbench}.
Various training-based methods have been proposed to improve spatial reasoning.
Some curate large-scale spatial data or design hierarchical curricula to strengthen basic spatial capabilities~\cite{chen2024spatialvlm,tang2024sparkle,sensenova-si}, while others inject geometric inductive biases such as depth features~\cite{cheng2024spatialrgpt} or employ reinforcement learning with spatial rewards~\cite{batra2025spatialthinker,Spatial-ssrl}.
However, these methods rely on additional training with specialized spatial data, limiting their generalizability across diverse spatial reasoning scenarios.

Beyond training-based methods, training-free methods enhance spatial reasoning without modifying the model.
Prompting-based methods decompose spatial tasks into structured subproblems or enrich input with spatial cues such as camera trajectories~\cite{li2025see}.
Tool-augmented methods construct 3D representations or simulate alternative viewpoints to enable explicit spatial reasoning~\cite{lee2025perspective,wang2026allocentric,yang2025mindjourney}.
However, these methods primarily focus on viewpoint-dependent reasoning such as reference frame transformation, overlooking other spatial hallucination scenarios such as perspective effects.
Moreover, they lack a unified framework to systematically address diverse spatial reasoning hallucinations.
Following this training-free paradigm, our proposed Geo3R addresses these limitations as a unified hallucination mitigation framework that covers perspective effects, object orientation, and viewpoint changes through multi-space geometric representations.

\subsection{Hallucination Mitigation in MLLMs}
Hallucination in MLLMs, where models generate content unfaithful to the visual input, remains a critical challenge~\cite{bai2024hallucination}.
Such hallucinations are commonly classified into existence, attribute, and relation categories~\cite{tu2025ode,trihe}.
Early research primarily focused on object hallucination, with various benchmarks and mitigation methods established~\cite{li2023evaluating,rohrbach2018object,huang2024opera,zhang2025mllms}.
Recently, attention has shifted toward relation hallucination, where models mischaracterize interactions between correctly identified entities.
Compared to object hallucination, relation hallucination is more prevalent and difficult to mitigate, as it requires reasoning about inter-object interactions rather than recognizing individual entities. 
For instance, Tri-HE~\cite{trihe} reports that the relation hallucination rate of LLaVA-1.5 reaches 24.8\%, nearly twice its object hallucination rate of 12.4\%.

To address relation hallucination, several mitigation methods have been proposed~\cite{trihe,zheng2025reefknot,chen2025spatial,wu2025mitigating}.
Tri-HE~\cite{trihe} decomposes model responses into structured triplets and performs self-alignment to correct relational hallucinations.
Reefknot~\cite{zheng2025reefknot} detects low-confidence relational predictions and recalibrates them.
AdaptVis~\cite{chen2025spatial} adaptively adjusts attention distributions to improve spatial relation understanding.
However, research on relation hallucination remains in its early stages, and effective mitigation is still an open challenge.

\section{Preliminary Analysis}
\label{sec:preliminary}
\begin{table}[t]
\caption{Accuracy (\%) of hallucination mitigation methods on spatial reasoning tasks.}
\label{tab:prelim}
\centering
\resizebox{0.95\columnwidth}{!}{
\begin{tabular}{lccccccc}
\toprule
Method & Height & Closer & Viewpoint & Facing & Left & Front & \textbf{Avg} \\
\midrule
Random & 50.00 & 50.00 & 25.00 & 50.00 & 50.00 & 50.00 & 45.83 \\
\midrule
LLaVA-1.5 & 51.40 & 58.60 & 24.80 & 41.20 & 40.20 & 50.60 & 44.47 \\
OPERA & 52.80 & 57.60 & 23.20 & 43.60 & 40.20 & 52.20 & 44.93\,{\tiny \textcolor{green!60!black}{+0.46}} \\
Tri-HE & 49.20 & 53.40 & 21.80 & 50.00 & 43.00 & 45.80 & 43.87\,{\tiny \textcolor{red}{-0.60}} \\
Reefknot & 51.00 & 58.40 & 25.40 & 39.40 & 39.00 & 49.60 & 43.80\,{\tiny \textcolor{red}{-0.67}} \\
AdaptVis & 50.60 & 58.60 & 23.40 & 57.80 & 41.40 & 44.60 & 46.07\,{\tiny \textcolor{green!60!black}{+1.60}} \\
\bottomrule
\end{tabular}
}
\end{table}

\begin{figure}[t]
\centering
\includegraphics[width=\columnwidth]{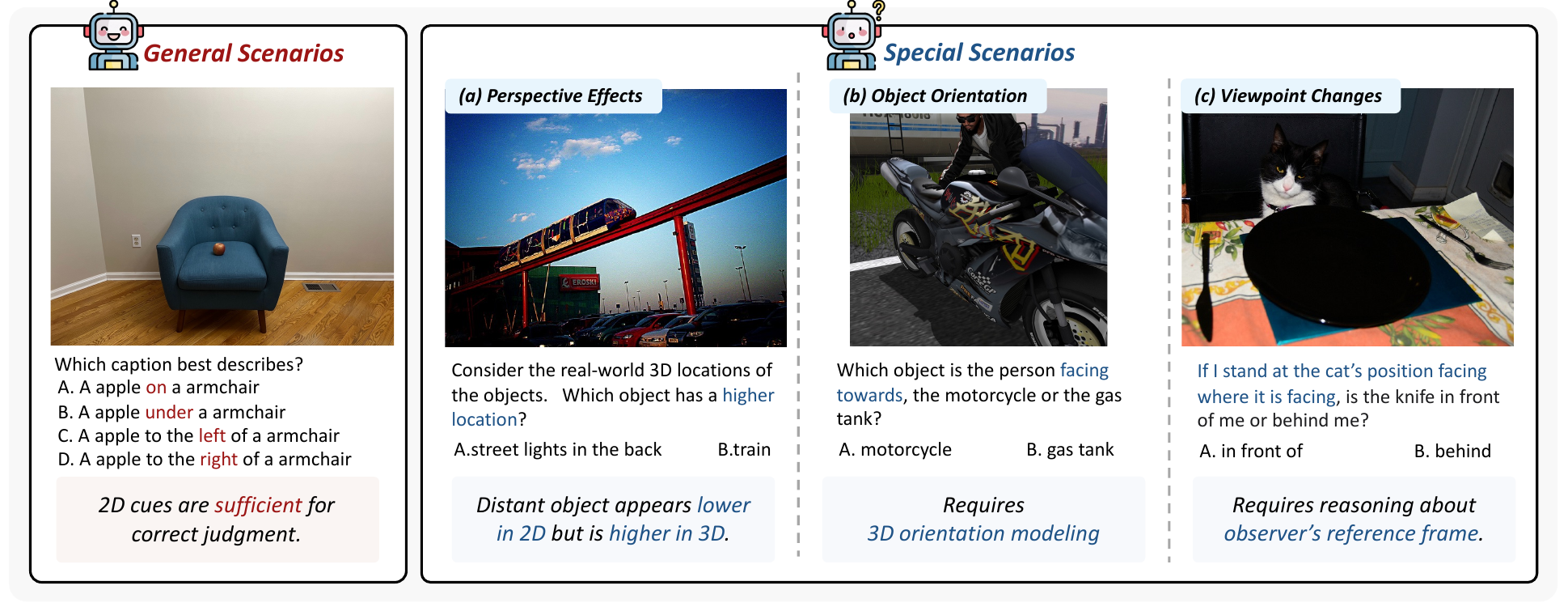}
\caption{General spatial relations (left) versus three spatial reasoning hallucination scenarios (a--c).}
\Description{xx}
\label{fig:case_analysis}
\end{figure}

While existing hallucination mitigation methods have shown promising results in reducing object and attribute hallucinations, their effectiveness on spatial reasoning tasks remains unclear.
To investigate, we apply one general and three relation hallucination mitigation methods~\cite{huang2024opera,trihe,zheng2025reefknot,chen2025spatial} to LLaVA-1.5 and evaluate on six representative spatial reasoning tasks from the 3DSRBench~\cite{ma20253dsrbench}, with 500 samples per task.

The results are summarized in Table~\ref{tab:prelim}.
LLaVA-1.5 itself performs poorly on spatial reasoning, with Viewpoint scoring only 24.80\%, below the 25\% random baseline, and Facing at 41.20\% and Left at 40.20\% also falling well below their 50\% random baselines.
Moreover, applying existing hallucination mitigation methods fails to address these deficiencies. Tri-HE and Reefknot degrade overall performance by 0.60\% and 0.67\% respectively, while OPERA yields only a marginal improvement of 0.46\%.
Although AdaptVis achieves the highest average gain of 1.60\%, this improvement is driven almost entirely by Facing, while performance on other tasks such as Viewpoint and Front declines.

To understand these results, we analyze the underlying causes.
The ineffectiveness of OPERA is understandable, as it primarily targets object hallucination.
However, the remaining three methods are specifically designed for relation or spatial hallucination, yet still fail to yield consistent improvements.
We thus further analyze the hallucination patterns and find that these failures concentrate in tasks that require understanding 3D spatial structure rather than surface-level visual semantics.
As illustrated in Figure~\ref{fig:case_analysis}, for general spatial relations where 2D image cues are sufficient, existing methods can mitigate model hallucinations effectively.
However, when 2D visual cues diverge from the true 3D spatial relationships, such as perspective effects, object orientation, and viewpoint changes, models rely on cues in the 2D image representation rather than reasoning about the underlying spatial reality.
This explains why existing hallucination mitigation methods fail on spatial reasoning tasks: they improve how models describe visually perceived entities and their interactions, but correct spatial judgments require reasoning about 3D structure beyond 2D pixels.

Based on this analysis, we define \textbf{spatial reasoning hallucination} as a subcategory of relation hallucination, arising from insufficient spatial structure modeling rather than visual-textual misalignment. 
Unlike general hallucinations where models fail to faithfully describe what they perceive, spatial reasoning hallucination occurs because 2D visual cues are inherently insufficient to determine the true 3D spatial configuration. 
Specifically, three typical scenarios of this type of hallucination are identified:

\textbf{Perspective effects.} Distortions introduced by projective transformation, including scale changes and foreshortening. 
Models may misjudge relative object sizes, distances, or spatial positions when perspective cues conflict with the 3D spatial reality.

\textbf{Object orientation.} Spatial judgments that depend on the facing direction of objects. 
Answering questions such as whether objects face each other or point in the same direction requires explicit orientation modeling beyond simple positional analysis.

\textbf{Viewpoint changes.} Spatial relationships that depend on the observer's viewpoint.
The same physical configuration appears differently from different positions, requiring models to reason about viewpoint and reference frame differences.

\section{Method}
\label{sec:method}

\begin{figure*}[t]
\centering
\includegraphics[width=0.95\textwidth]{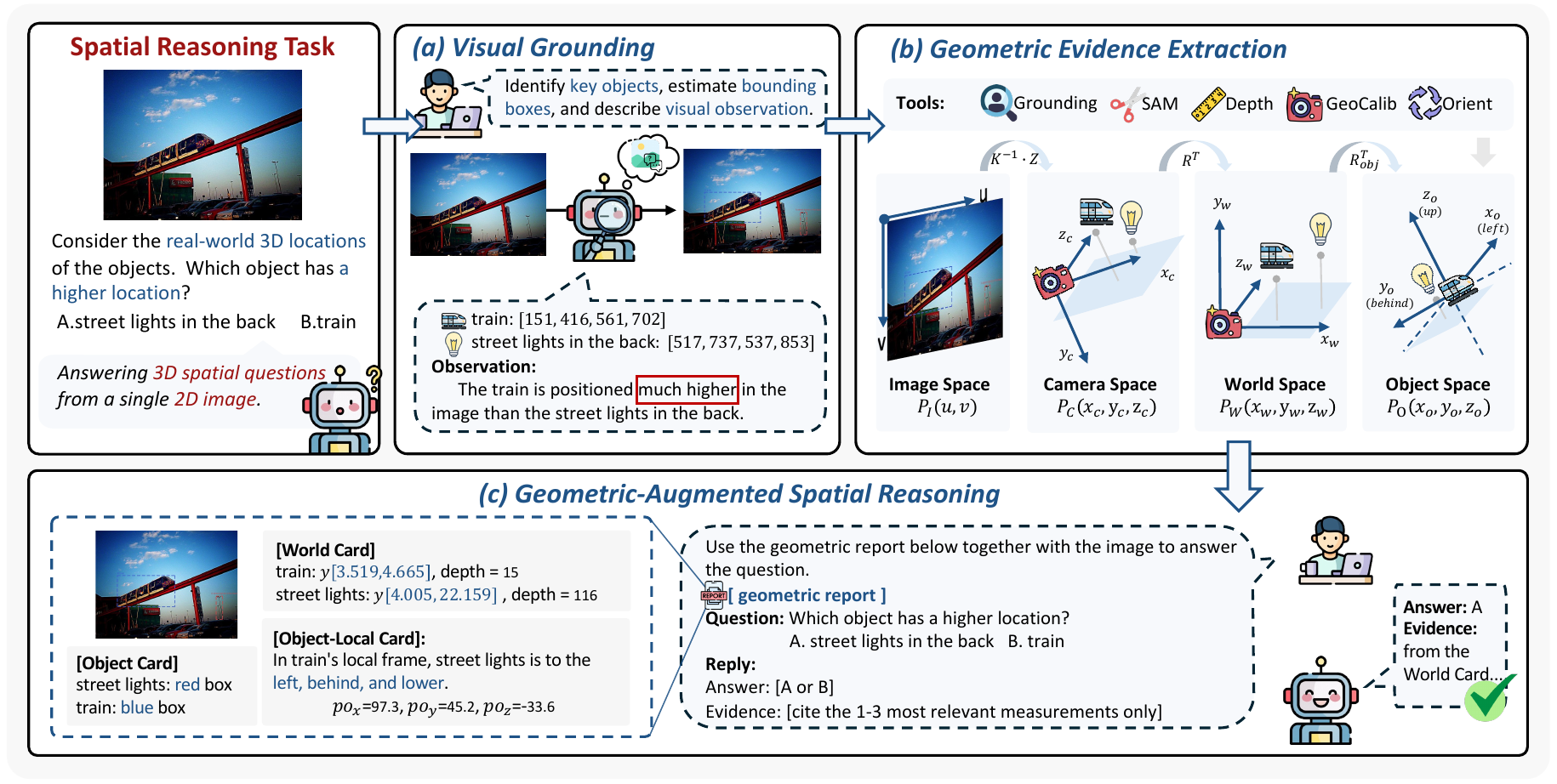}
\caption{Overview of Geo3R.}
\Description{}
\label{fig:overview}
\end{figure*}
As illustrated in Figure~\ref{fig:overview}, Geo3R is a training-free, plug-and-play framework that augments any MLLM with explicit geometric reasoning from a single image.
The core idea is to progressively lift 2D pixel observations into multi-space geometric evidence through four coordinate frames: image space $\mathbf{P}_I$, camera space $\mathbf{P}_C$, gravity-aligned world space $\mathbf{P}_W$, and object-local space $\mathbf{P}_O$, jointly addressing all three spatial hallucination scenarios identified in Section~\ref{sec:preliminary}.
To this end, Geo3R operates in three sequential stages:
(a)~visual grounding, which identifies task-relevant objects and their image regions,
(b)~geometric evidence extraction, which constructs multi-space 3D representations using geometric estimation modules,
and (c)~geometric-augmented spatial reasoning, which organizes the extracted evidence into structured cards to guide the MLLM's final answer.

\subsection{Visual Grounding}
\label{sec:visual_grounding}

Given an image $I$ and a spatial reasoning question $Q$ with answer choices $\{c_1,\ldots,c_m\}$, the visual grounding stage extracts the minimal set of task-relevant objects and their image regions, producing three outputs: a set of object names $\mathcal{O}=\{o_1,\ldots,o_n\}$, bounding boxes $\mathcal{B}$ in pixel coordinates, and a visual observation $s$ that captures the MLLM's initial perception of the scene based on 2D visual cues.
Concretely, the MLLM takes the image along with the question and answer choices to jointly extract the object names $\mathcal{O}$, their bounding boxes $\mathcal{B}$, and the visual observation $s$. For MLLMs with limited grounding capabilities, GroundingDINO~\cite{liu2024grounding} is used to obtain the bounding boxes.

\subsection{Geometric Evidence Extraction}
\label{sec:geo_extraction}

This stage constructs a multi-space geometric representation by integrating four coordinate frames through a progressive transformation chain, lifting 2D pixel observations into structured geometric evidence.

\subsubsection{Coordinate System Definitions.}
The multi-space geometric representation is built upon four coordinate spaces defined in distinct reference frames.

\textbf{Image space} $\mathbf{P}_I\!\in\!\mathbb{R}^2$: 2D pixel coordinates in the image plane.

\textbf{Camera space} $\mathbf{P}_C\!\in\!\mathbb{R}^3$: 3D coordinates centered at the camera.

\textbf{World space} $\mathbf{P}_W\!\in\!\mathbb{R}^3$: obtained by rotating camera space to align with the gravity direction.

\textbf{Object-local space} $\mathbf{P}_O\!\in\!\mathbb{R}^3$: centered at the object centroid, with axes aligned to the object's intrinsic orientation.

\subsubsection{Back-Projection to Camera Space.}
This stage lifts 2D pixel observations into 3D camera-space coordinates by combining monocular depth estimation with camera intrinsic calibration.

Specifically, DepthPro~\cite{bochkovskii2024depth} produces a per-pixel metric depth map $D$, from which the depth $z_i$ for each object $o_i$ is computed.
GeoCalib~\cite{veicht2024geocalib} estimates the focal length $f$, from which we construct the intrinsic matrix $K$:
\begin{equation}
\label{eq:K}
K = \begin{pmatrix} f & 0 & c_x \\ 0 & f & c_y \\ 0 & 0 & 1 \end{pmatrix},
\end{equation}
where $(c_x, c_y)$ is the image center. Combined with segmentation masks refined by SAM~\cite{kirillov2023segment}, each pixel $(u,v)$ with depth $z$ is back-projected to camera space:
\begin{equation}
\label{eq:backproj}
\mathbf{P}_C = z \cdot K^{-1}
\begin{pmatrix} u \\ v \\ 1 \end{pmatrix}
= \begin{pmatrix} (u - c_x)\,z\,/\,f \\[2pt] (v - c_y)\,z\,/\,f \\[2pt] z \end{pmatrix}.
\end{equation}

\subsubsection{Gravity-Aligned World Transformation.}
This stage transforms camera-space coordinates into a gravity-aligned world frame, enabling physically meaningful height and depth comparisons.

GeoCalib additionally estimates the camera's orientation relative to gravity, parameterized by roll $\rho$ and pitch $\theta$. We construct the rotation matrix $R$ as:
\begin{equation}
\label{eq:R}
R = R_z(\!-\!\rho)\;R_x(\!-\!\theta),
\end{equation}
where $R_x$ and $R_z$ denote elementary rotation matrices about the $X$ and $Z$ axes, respectively.
Camera-space points are then transformed into the gravity-aligned world frame as $\mathbf{P}_W$:
\begin{equation}
\label{eq:cam2world}
\mathbf{P}_W = R^\top \mathbf{P}_C,
\end{equation}
in which $Y_W$ points upward against gravity.

For each object, we project all mask pixels into world coordinates and compute trimmed quantile ranges (10th--90th percentile) along each axis, yielding a 3D bounding extent $\mathcal{E}_i$:
\begin{equation}
\label{eq:extent}
\mathcal{E}_i = \bigl\{[x_{\min},x_{\max}],\;[y_{\min},y_{\max}],\;[z_{\min},z_{\max}]\bigr\}.
\end{equation}

\subsubsection{Object-Local Transformation.}
This stage estimates object orientations and expresses relative object positions in object-local coordinate frames.

We use Orient Anything V2~\cite{wang2026orient} to estimate each object's 3D orientation, obtaining a rotation matrix $R_{\text{obj}}^C$ that encodes the object's local axes in camera space. 
This is then converted to the world frame as $R_{\text{obj}}^W$:
\begin{equation}
\label{eq:Robjworld}
R_{\text{obj}}^W = R^\top R_{\text{obj}}^C.
\end{equation}
Given a reference object with world-space centroid $\mathbf{c}_{\text{ref}}$ and orientation $R_{\text{ref}}^W$, the position of a target object in the reference's local frame is computed as:
\begin{equation}
\label{eq:world2obj}
\mathbf{P}_O = (R_{\text{ref}}^W)^\top\,(\mathbf{P}_{W,\text{tgt}} - \mathbf{c}_{\text{ref}}).
\end{equation}

\subsubsection{Per-Object Profile and Pairwise Relation Aggregation.}
This stage aggregates geometric measurements into per-object profiles and pairwise relations.

For each grounded object $o_i$, we compile a unified geometric profile:
\begin{equation}
\label{eq:profile}
\mathcal{G}_i = \bigl\{\mathbf{P}_{C,i},\;\mathbf{P}_{W,i},\;z_i,\;\mathcal{E}_i,\;\alpha_i,\;v_i\bigr\},
\end{equation}
where $v_i$ denotes the camera-visible side derived from the estimated azimuth $\alpha_i$.

For each ordered pair $(o_i, o_j)$, we derive pairwise relations from two complementary coordinate frames. World-frame relations capture global spatial structure, including 3D distance, vertical and depth orderings, camera-relative direction, and heading relation between objects.

Object-local relations capture spatial structure relative to a reference object. 
From the target's local position $\mathbf{P}_O$, we derive the local offset, quantized direction, and the side of the reference facing the target.
When multiple targets share the same reference, a comparative summary is additionally derived.

\subsection{Geometric-Augmented Spatial Reasoning}
\label{sec:geo_reasoning}

This stage organizes the geometric evidence into structured cards and integrates them into the MLLM prompt to guide the MLLM's spatial reasoning.

The card-based design is motivated by two considerations.
First, different spatial reasoning scenarios require evidence from different coordinate frames.
Second, presenting all raw measurements in an unstructured form risks overwhelming the MLLM, leading to selection of irrelevant or contradictory cues.
We therefore organize the extracted geometric evidence into three structured cards that decompose evidence by coordinate frame, each accompanied by field notes that explain the underlying conventions.
Importantly, all cards are task-agnostic, meaning the same set of fields is provided for every question, avoiding the need for a task-type classifier and allowing the MLLM to flexibly combine evidence across cards.
Furthermore, each field pairs a continuous measurement with a discrete human-readable summary, enabling the MLLM to verify reasoning against raw values while using the summaries as interpretive cues.

\textbf{Object Card.} This card maps each detected object to a color-coded bounding box drawn on the annotated image, grounding textual object names to visual regions and establishing a consistent reference across all cards.

\textbf{World Card.} For each object, this card reports the 3D bounding extent $\mathcal{E}_i$ in the gravity-aligned world frame, the camera depth $z_i$, and the camera-visible side $v_i$.
For each object pair, it provides the vertical ordering, depth ordering, 3D distance $d_{ij}$, heading relation $h_{ij}$, and camera-relative direction.

\textbf{Object-Local Card.} For each reference--target pair, this card reports the local offset vector $\mathbf{P}_O$, the quantized local direction, the side of the reference facing the target, and the combined vertical-topdown spatial relation.
When multiple targets share the same reference object, a comparative summary across the three local axes is included.

All three cards, along with usage guidelines, are provided in the second-stage prompt. The MLLM then synthesizes the geometric evidence with its visual understanding to produce the final answer. Full prompt templates, card examples, and method details are provided in the appendix.

\begin{table*}[t]
\caption{Comparison with MLLMs on spatial reasoning benchmarks. Best \colorbox{hlblue_light}{highlighted}, second best \underline{underlined}.
$^V$\,VSB, $^S$\,3DSR, $^C$\,CVB.}
\label{tab:main_results}
\centering
\resizebox{\textwidth}{!}{
\begin{tabular}{l c c c c c c c c c c c c c c c c c c c}
\toprule
 & \multicolumn{7}{c}{Perspective} & \multicolumn{6}{c}{Orientation} & \multicolumn{5}{c}{Viewpoint} &  \\
\cmidrule(lr){2-8} \cmidrule(lr){9-14} \cmidrule(lr){15-19}
Tasks & \shortstack{Loc.$^{S}$\\Above} & \shortstack{Hgt.$^{S}$\\Higher} & \shortstack{Loc.$^{S}$\\Cam.} & \shortstack{Loc.$^{S}$\\Next.} & \shortstack{Loc.$^{S}$\\Obj.} & \shortstack{Depth$^{C}$\\~} & \shortstack{Dist.$^{C}$\\~} & \shortstack{Cam.$^{V}$\\Orient.} & \shortstack{Psn.$^{V}$\\Orient.} & \shortstack{Orient.$^{S}$\\View.} & \shortstack{Orient.$^{S}$\\Facing} & \shortstack{Orient.$^{S}$\\Same-Dir.} & \shortstack{Orient.$^{S}$\\Parallel} & \shortstack{Orient.$^{S}$\\Twd.} & \shortstack{Orient.$^{S}$\\Left} & \shortstack{Orient.$^{S}$\\Front} & \shortstack{Psn.$^{V}$\\Rel-Dir.} & \shortstack{Cam.$^{V}$\\Rel-Dir.} & \textbf{Avg} \\
\midrule
Random & 50.00 & 50.00 & 50.00 & 50.00 & 50.00 & 50.00 & 50.00 & 25.00 & 25.00 & 25.00 & 50.00 & 50.00 & 50.00 & 25.00 & 50.00 & 50.00 & 25.00 & 25.00 & 41.67 \\
\midrule
\rowcolor{hlblue} \multicolumn{20}{l}{\textit{General-Purpose MLLMs}} \\
LLaVA-1.5 & 56.72 & 50.51 & 58.92 & 55.16 & 47.43 & 51.00 & 48.83 & 10.44 & 51.61 & 24.85 & 42.63 & 45.78 & 50.74 & 26.02 & 40.54 & 51.74 & 37.53 & 27.69 & 43.23 \\
LLaVA-OV & 64.23 & 59.28 & 74.12 & 55.16 & 53.86 & 51.67 & 54.17 & 23.29 & 36.24 & 37.54 & 60.98 & 50.87 & 52.21 & 24.85 & 38.83 & 56.25 & 32.54 & 40.78 & 48.16 \\
InternVL3 & 63.80 & 56.67 & 85.03 & 57.37 & 61.71 & 86.00 & 71.17 & 26.41 & 38.35 & 42.57 & 65.32 & 52.91 & 57.37 & 25.07 & 34.53 & 63.95 & 35.87 & 51.95 & 54.22 \\
Qwen3-VL & 69.36 & 54.86 & 87.68 & 58.41 & 64.71 & \cellcolor{hlblue_light}95.00 & 47.67 & 28.61 & 45.08 & 42.93 & 68.79 & 55.81 & 56.34 & 27.04 & 35.10 & 63.95 & 36.58 & 53.41 & 55.07 \\
Kimi-VL & \underline{70.95} & 59.64 & 78.02 & 73.89 & 63.86 & 84.83 & 64.83 & 27.21 & 56.43 & 40.82 & 67.34 & 49.27 & 56.78 & 30.76 & 34.81 & 62.94 & 43.94 & 49.52 & 56.44 \\
Gemini-3-Flash & 70.09 & 53.26 & 80.90 & 73.60 & 68.43 & 91.67 & 88.50 & \underline{35.54} & 42.57 & 57.14 & 73.55 & \underline{65.84} & \cellcolor{hlblue_light}69.17 & 42.64 & \underline{62.46} & \cellcolor{hlblue_light}82.41 & 57.36 & \underline{60.97} & 65.34 \\
GPT-5 & \cellcolor{hlblue_light}72.25 & \cellcolor{hlblue_light}72.90 & 87.17 & 74.63 & \underline{77.57} & 92.33 & \underline{90.67} & 30.02 & 41.37 & 54.59 & 70.23 & 62.79 & 65.49 & 40.67 & 56.59 & \underline{80.96} & 52.97 & \cellcolor{hlblue_light}63.45 & 65.93 \\
\midrule
\rowcolor{hlblue} \multicolumn{20}{l}{\textit{Spatial Understanding Models}} \\
SpaceLLaVA & 52.82 & 51.01 & 50.44 & 52.06 & 49.43 & 50.50 & 48.67 & 15.46 & 46.08 & 25.51 & 52.31 & 45.78 & 50.15 & 24.49 & 50.43 & 50.73 & 35.75 & 28.99 & 43.37 \\
SpatialRGPT & 59.54 & 59.28 & 70.35 & 68.14 & 58.00 & 77.17 & 61.33 & 27.21 & 38.15 & 34.91 & 56.07 & 54.65 & 53.39 & 24.56 & 39.54 & 58.14 & 33.85 & 41.57 & 50.88 \\
Spatial-SSRL & 57.66 & 53.33 & 82.23 & 61.36 & 64.14 & 90.00 & 76.83 & 29.62 & 41.87 & 37.68 & 64.16 & 52.62 & 54.72 & 21.57 & 34.67 & 61.92 & 38.84 & 42.81 & 53.67 \\
SenseNova-SI & 52.31 & 53.12 & 81.86 & 52.21 & 64.86 & 85.17 & 82.67 & 22.29 & \underline{57.43} & 30.10 & \underline{73.84} & 53.63 & 55.46 & 22.67 & 34.67 & 55.96 & 42.76 & 46.31 & 53.74 \\
SpatialThinker & 66.04 & 59.64 & 75.96 & 66.22 & 59.71 & 85.50 & 79.17 & 30.02 & 50.60 & 43.00 & 69.65 & 50.73 & 53.69 & 22.89 & 35.10 & 62.21 & 38.36 & 44.22 & 55.15 \\
\midrule
\rowcolor{hlblue} \multicolumn{20}{l}{\textit{Ours}} \\
+ LLaVA-1.5 & 56.58 & 51.23 & 55.09 & 56.78 & 45.71 & 56.67 & 51.17 & 10.64 & \cellcolor{hlblue_light}67.77 & 29.59 & 46.82 & 53.20 & 52.95 & 25.87 & 49.57 & 52.91 & 35.99 & 31.19 & 46.10 \\
+ Qwen3-VL & 69.51 & 71.30 & \cellcolor{hlblue_light}88.42 & \cellcolor{hlblue_light}84.96 & 74.14 & 90.83 & 70.17 & 32.53 & 55.52 & \underline{64.87} & 70.66 & 63.52 & 67.11 & \underline{45.19} & 45.70 & 72.24 & \underline{60.81} & 60.01 & \underline{65.97} \\
+ Gemini-3-Flash & 69.15 & \underline{71.96} & \underline{88.05} & \underline{80.38} & \cellcolor{hlblue_light}85.71 & \underline{93.83} & \cellcolor{hlblue_light}92.83 & \cellcolor{hlblue_light}39.26 & 43.78 & \cellcolor{hlblue_light}72.23 & \cellcolor{hlblue_light}77.75 & \cellcolor{hlblue_light}67.01 & \underline{67.99} & \cellcolor{hlblue_light}65.01 & \cellcolor{hlblue_light}78.94 & 79.65 & \cellcolor{hlblue_light}74.47 & 55.27 & \cellcolor{hlblue_light}72.40 \\
\bottomrule
\end{tabular}
}
\end{table*}

\section{Experiments}
\label{sec:experiments}

\subsection{Experimental Setup}
\label{sec:setup}

\subsubsection{Datasets and Tasks}
\label{sec:datasets}

We evaluate Geo3R on 18 spatial reasoning tasks totaling 17{,}493 samples across three benchmarks.

Following the three scenarios defined in Section~\ref{sec:preliminary}, these tasks are grouped into Perspective, Orientation, and Viewpoint.
Perspective covers 7 tasks with 6{,}698 samples, Orientation covers 6 tasks with 5{,}422 samples, and Viewpoint covers 5 tasks with 5{,}373 samples.

\textbf{ViewSpatial-Bench (VSB)}~\cite{li2025viewspatial} is a multi-perspective spatial reasoning benchmark that probes both camera-perspective and person-perspective understanding.
We evaluate on 4 tasks spanning the Orientation and Viewpoint scenarios.

\textbf{3DSRBench (3DSR)}~\cite{ma20253dsrbench} is a comprehensive 3D spatial reasoning benchmark constructed from real-world images.
We evaluate on 12 tasks covering all three scenarios, including location and height comparisons, orientation judgments, and direction reasoning.

\textbf{CV-Bench (CVB)}~\cite{tong2024cambriancvbench} is a vision-centric benchmark introduced in Cambrian-1 that includes 2D and 3D subsets. We evaluate on the 3D subset with 2 Perspective tasks, depth ordering and distance estimation, that require reliable depth understanding.

\subsubsection{Baselines}
\label{sec:baselines}

We compare against three categories of existing methods.

\textbf{General-purpose MLLMs.} We include seven MLLMs spanning open-source and proprietary models of varying scales, including LLaVA-1.5-7B~\cite{llava-1.5}, LLaVA-OV-7B~\cite{llava-onevision}, InternVL3-8B~\cite{zhu2025internvl3}, Qwen3-VL-8B~\cite{yang2025qwen3}, Kimi-VL-A3B~\cite{team2025kimi}, Gemini-3-Flash~\cite{team2023gemini}, and GPT-5~\cite{singh2025openai}. These models are evaluated in a zero-shot setting without any spatial-specific adaptation.

\textbf{Spatial understanding models.} We include five models explicitly trained or fine-tuned for spatial reasoning, including Space\-LLaVA-7B~\cite{chen2024spatialvlm}, SpatialRGPT-8B~\cite{cheng2024spatialrgpt}, Spatial-SSRL-7B~\cite{Spatial-ssrl}, SenseNova-SI-7B~\cite{sensenova-si}, and SpatialThinker-7B~\cite{batra2025spatialthinker}. These models incorporate spatial-aware training data or learning paradigms to enhance 3D understanding.

\textbf{Hallucination mitigation and tool-augmented methods.} We include one general hallucination mitigation method, OPERA~\cite{huang2024opera}, and three relation hallucination methods, Tri-HE~\cite{trihe}, Reefknot~\cite{zheng2025reefknot}, and AdaptVis~\cite{chen2025spatial}. Most of these methods require access to model internals such as attention weights or hidden states, limiting their compatibility to specific architectures. For a fair and unified comparison, all four methods are evaluated on LLaVA-1.5-7B, the only model supported by all their official implementations. We also compare with APC-VLM~\cite{lee2025perspective}, a training-free tool-augmented method, as other spatial reasoning methods either require additional training, target different tasks, or are not publicly available. APC-VLM is evaluated on Qwen3-VL under the same protocol as Geo3R.

\subsubsection{Implementation Details}
\label{sec:implementation}

All three benchmarks are multiple-choice in our evaluated subsets.
For all experiments, we use the official evaluation splits as the ground truth. 
Following standard protocols, we report Accuracy based on exact matching between the predicted and the ground-truth answers.
We report the simple average across all 18 tasks as the overall metric.
Detailed task descriptions, scenario mappings, and implementation details are provided in the appendix.

\begin{table*}[t]
\caption{Comparison with hallucination mitigation methods on LLaVA-1.5 (best \colorbox{hlblue_light}{highlighted}, second best \underline{underlined}), and the effect of adding Geo3R to Qwen3-VL and Gemini-3-Flash.}
\label{tab:ablation_merged}
\centering
\resizebox{\textwidth}{!}{
\begin{tabular}{l c c c c c c c c c c c c c c c c c c c}
\toprule
 & \multicolumn{7}{c}{Perspective} & \multicolumn{6}{c}{Orientation} & \multicolumn{5}{c}{Viewpoint} &  \\
\cmidrule(lr){2-8} \cmidrule(lr){9-14} \cmidrule(lr){15-19}
Tasks & \shortstack{Loc.$^{S}$\\Above} & \shortstack{Hgt.$^{S}$\\Higher} & \shortstack{Loc.$^{S}$\\Cam.} & \shortstack{Loc.$^{S}$\\Next.} & \shortstack{Loc.$^{S}$\\Obj.} & \shortstack{Depth$^{C}$\\~} & \shortstack{Dist.$^{C}$\\~} & \shortstack{Cam.$^{V}$\\Orient.} & \shortstack{Psn.$^{V}$\\Orient.} & \shortstack{Orient.$^{S}$\\View.} & \shortstack{Orient.$^{S}$\\Facing} & \shortstack{Orient.$^{S}$\\Same-Dir.} & \shortstack{Orient.$^{S}$\\Parallel} & \shortstack{Orient.$^{S}$\\Twd.} & \shortstack{Orient.$^{S}$\\Left} & \shortstack{Orient.$^{S}$\\Front} & \shortstack{Psn.$^{V}$\\Rel-Dir.} & \shortstack{Cam.$^{V}$\\Rel-Dir.} & \textbf{Avg} \\
\midrule
Random & 50.00 & 50.00 & 50.00 & 50.00 & 50.00 & 50.00 & 50.00 & 25.00 & 25.00 & 25.00 & 50.00 & 50.00 & 50.00 & 25.00 & 50.00 & 50.00 & 25.00 & 25.00 & 41.67 \\
\midrule
\rowcolor{blockbg} LLaVA-1.5 & \underline{56.72} & 50.51 & \underline{58.92} & 55.16 & 47.43 & 51.00 & 48.83 & 10.44 & 51.61 & 24.85 & 42.63 & 45.78 & \underline{50.74} & \underline{26.02} & 40.54 & 51.74 & \underline{37.53} & 27.69 & 43.23 \\
\rowcolor{blockbg} + OPERA & 56.00 & \underline{51.09} & 58.55 & 52.65 & 48.14 & 56.17 & 48.17 & 10.04 & 51.10 & 24.93 & 44.65 & 46.51 & 48.38 & 25.87 & 40.69 & \underline{52.18} & 37.17 & \underline{29.55} & 43.44\,{\tiny \textcolor{green!60!black}{+0.21}} \\
\rowcolor{blockbg} + Tri-HE & 55.35 & 48.99 & 56.19 & 51.77 & \cellcolor{hlblue_light}50.00 & 51.67 & 48.33 & 8.94 & 56.53 & 24.78 & \underline{50.29} & 46.22 & 49.56 & 24.78 & \underline{42.84} & 46.66 & 33.61 & 27.98 & 43.03\,{\tiny \textcolor{red}{-0.20}} \\
\rowcolor{blockbg} + Reefknot & \cellcolor{hlblue_light}57.08 & 50.14 & \cellcolor{hlblue_light}59.51 & \underline{55.46} & \underline{49.14} & 52.50 & 46.67 & \underline{10.54} & 50.00 & \underline{25.29} & 41.76 & 46.80 & 48.67 & 25.58 & 39.83 & 51.16 & \cellcolor{hlblue_light}37.77 & 29.44 & 43.19\,{\tiny \textcolor{red}{-0.04}} \\
\rowcolor{blockbg} + AdaptVis & 46.46 & 49.49 & 55.53 & 55.01 & 48.71 & \cellcolor{hlblue_light}60.50 & \underline{50.50} & 9.14 & \underline{59.24} & 23.54 & \cellcolor{hlblue_light}57.23 & \underline{48.55} & 50.59 & \cellcolor{hlblue_light}27.26 & 41.83 & 45.49 & 35.04 & 26.45 & \underline{43.92}\,{\tiny \textcolor{green!60!black}{+0.69}} \\
\rowcolor{blockbg} \textbf{+ Geo3R} & 56.58 & \cellcolor{hlblue_light}51.23 & 55.09 & \cellcolor{hlblue_light}56.78 & 45.71 & \underline{56.67} & \cellcolor{hlblue_light}51.17 & \cellcolor{hlblue_light}10.64 & \cellcolor{hlblue_light}67.77 & \cellcolor{hlblue_light}29.59 & 46.82 & \cellcolor{hlblue_light}53.20 & \cellcolor{hlblue_light}52.95 & 25.87 & \cellcolor{hlblue_light}49.57 & \cellcolor{hlblue_light}52.91 & 35.99 & \cellcolor{hlblue_light}31.19 & \cellcolor{hlblue_light}46.10\,{\tiny \textcolor{green!60!black}{+2.87}} \\
\midrule
Qwen3-VL & 69.36 & 54.86 & 87.68 & 58.41 & 64.71 & 95.00 & 47.67 & 28.61 & 45.08 & 42.93 & 68.79 & 55.81 & 56.34 & 27.04 & 35.10 & 63.95 & 36.58 & 53.41 & 55.07 \\
+ APC-VLM & 49.42 & 45.43 & 48.97 & 41.30 & 53.00 & 74.17 & 61.17 & 22.49 & 45.88 & 22.45 & 61.42 & 51.45 & 51.33 & 22.81 & 53.44 & 45.93 & 40.97 & 57.08 & 47.15\,{\tiny \textcolor{red}{-7.92}} \\
\textbf{+ Geo3R} & 69.51 & 71.30 & 88.42 & 84.96 & 74.14 & 90.83 & 70.17 & 32.53 & 55.52 & 64.87 & 70.66 & 63.52 & 67.11 & 45.19 & 45.70 & 72.24 & 60.81 & 60.01 & 65.97\,{\tiny \textcolor{green!60!black}{+10.9}} \\
\midrule
Gemini-3-Flash & 70.09 & 53.26 & 80.90 & 73.60 & 68.43 & 91.67 & 88.50 & 35.54 & 42.57 & 57.14 & 73.55 & 65.84 & 69.17 & 42.64 & 62.46 & 82.41 & 57.36 & 60.97 & 65.34 \\
\textbf{+ Geo3R} & 69.15 & 71.96 & 88.05 & 80.38 & 85.71 & 93.83 & 92.83 & 39.26 & 43.78 & 72.23 & 77.75 & 67.01 & 67.99 & 65.01 & 78.94 & 79.65 & 74.47 & 55.27 & 72.40\,{\tiny \textcolor{green!60!black}{+7.06}} \\
\bottomrule
\end{tabular}
}
\end{table*}

\begin{figure}[t]
\centering
\includegraphics[width=\columnwidth]{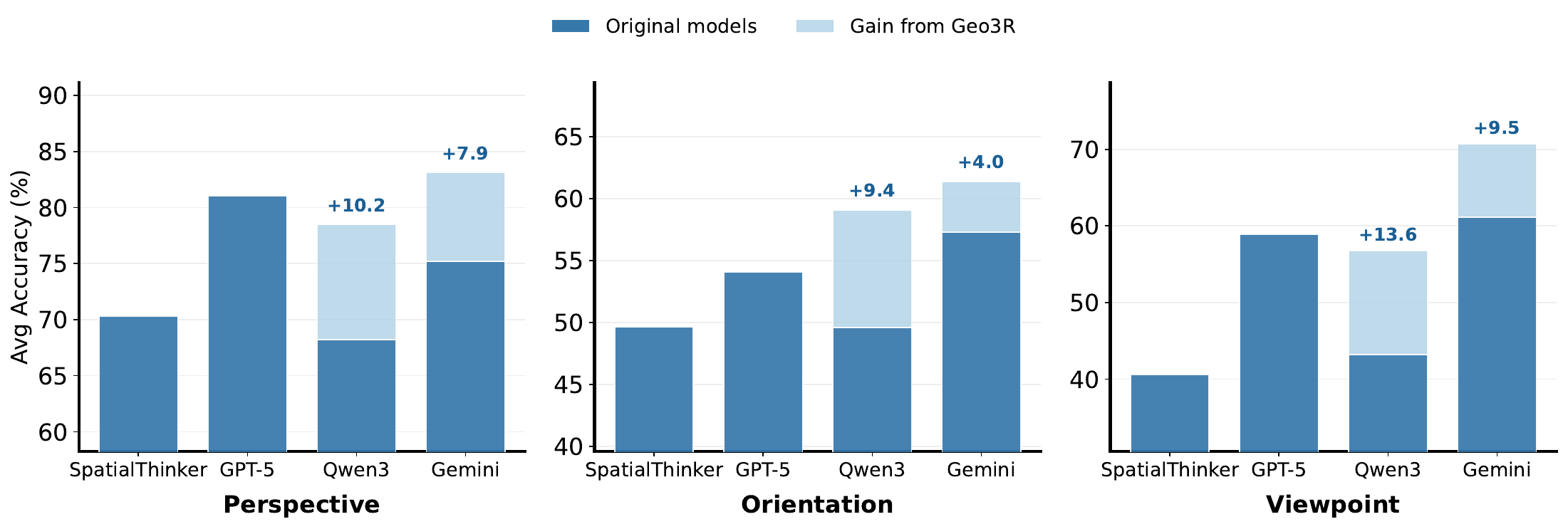}
\caption{Scenario-level average accuracy comparison across MLLMs, spatial understanding models, and Geo3R.}
\Description{}
\label{fig:bar_main}
\end{figure}
\subsection{Comparison with MLLMs}

Table~\ref{tab:main_results} presents the performance of all models across 18 spatial reasoning tasks.

\textbf{General-purpose MLLMs lack robust spatial reasoning.}
These models show limited spatial reasoning across all three scenarios, with accuracy declining sharply from Perspective to Orientation and Viewpoint.
On Perspective tasks, even stronger models show inconsistent performance, with GPT-5 and Gemini-3-Flash reaching 92.33\% and 91.67\% on depth ordering but dropping to 72.90\% and 53.26\% on height comparison.
Weaker models such as LLaVA-1.5 fare worse, hovering near the 50\% random baseline on most Perspective tasks.
The situation worsens on Orientation and Viewpoint tasks, which tend to involve more complex 3D reasoning.
GPT-5 achieves only 40.67\% on towards object and Gemini-3-Flash only 42.64\%. Weaker models fall further below chance, with LLaVA-1.5 scoring only 10.44\% on camera orientation, well below the 25\% random baseline.
These results highlight that spatial reasoning, particularly orientation and viewpoint understanding, remains a significant challenge for current MLLMs.

\textbf{Spatial understanding models show limited and inconsistent improvements.}
Models designed for spatial understanding improve on certain tasks but fail to generalize across all tasks.
Space\-LLaVA, despite being fine-tuned on synthetic spatial reasoning data, barely exceeds the random baseline on most tasks, with an overall average of 43.37\%. 
SenseNova-SI and SpatialThinker achieve strong results on selected Perspective and Orientation tasks, with SenseNova-SI reaching 82.67\% on distance estimation and SpatialThinker 85.50\% on depth ordering, yet both drop sharply on Viewpoint tasks, scoring only 22.67\% and 22.89\% on towards object, respectively.
These results indicate that existing spatial understanding models tend to improve specific spatial skills such as depth estimation while leaving others, particularly viewpoint reasoning, largely unaddressed.

\textbf{Geo3R improves spatial reasoning across base models.}
Geo3R with Gemini-3-Flash achieves the highest overall average of 72.40\%, surpassing GPT-5 by 6.47\% and obtaining the best score on 9 out of 18 individual tasks.
The gains are most notable on tasks where baselines struggle.
For example, orientation viewpoint improves from 57.14\% to 72.23\%, towards object from 42.64\% to 65.01\%, and orientation left from 62.46\% to 78.94\%.
Notably, Geo3R with Qwen3-VL-8B reaches an overall average of 65.97\%, surpassing GPT-5 at 65.93\% on the 18-task average despite the substantial difference in model scale. 
Similar task-level gains are observed, with orientation viewpoint rising from 42.93\% to 64.87\% and towards object from 27.04\% to 45.19\%.
Even on LLaVA-1.5, Geo3R improves the average from 43.23\% to 46.10\%.
Since Qwen3-VL-8B is comparable in scale to most models evaluated above, these consistent gains demonstrate that geometric augmentation is more effective than scaling model size or spatial-specific fine-tuning, while remaining applicable to diverse base architectures.

Figure~\ref{fig:bar_main} visualizes the scenario-level comparison.
All three scenarios benefit from Geo3R across all base models. On Qwen3-VL, Perspective improves by 10.24\%, Orientation by 9.45\%, and Viewpoint by 13.57\%. On Gemini-3-Flash, the corresponding improvements are 7.92\%, 4.04\%, and 9.50\%. The largest gains appear on Viewpoint, where baseline accuracy is lower and more room for improvement exists.

\begin{figure*}[t]
\centering
\includegraphics[width=0.95\textwidth]{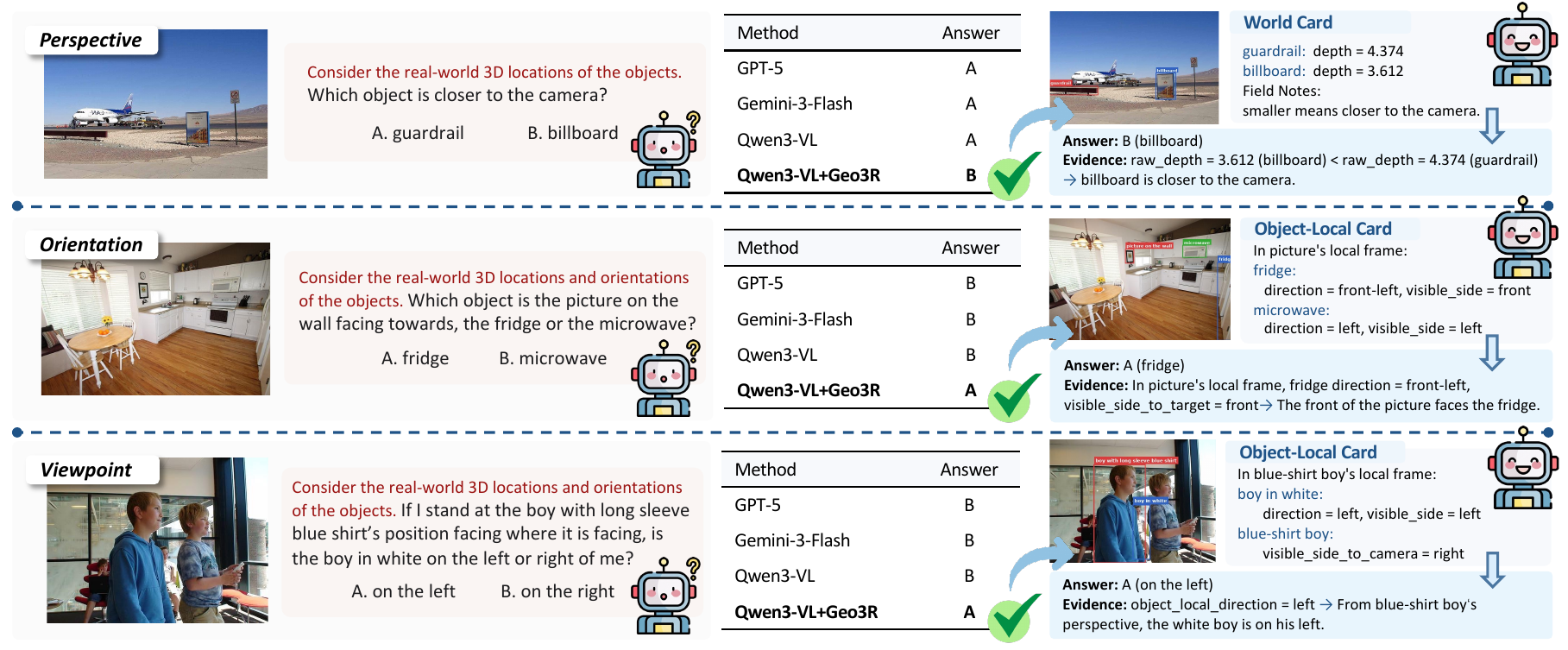}
\caption{Qualitative examples of Geo3R mitigating spatial reasoning hallucinations across the three scenarios.}
\Description{Three qualitative case studies showing how Geo3R mitigates spatial reasoning hallucinations across perspective, orientation, and viewpoint scenarios.}
\label{fig:cases}

\end{figure*}

Figure~\ref{fig:cases} presents qualitative examples illustrating how Geo3R mitigates spatial reasoning hallucinations across the three scenarios. In the perspective case, baselines misjudge object depth from 2D position, whereas the World Card provides metric depth for correct judgment. In the orientation case, the facing target is misidentified, but the Object-Local Card reveals the correct facing direction. In the viewpoint case, the relation is answered from the camera's perspective, while the Object-Local Card resolves it in the reference object's own frame.

\subsection{Comparison with Other Methods}
\begin{figure}[t]
\centering
\includegraphics[width=\columnwidth]{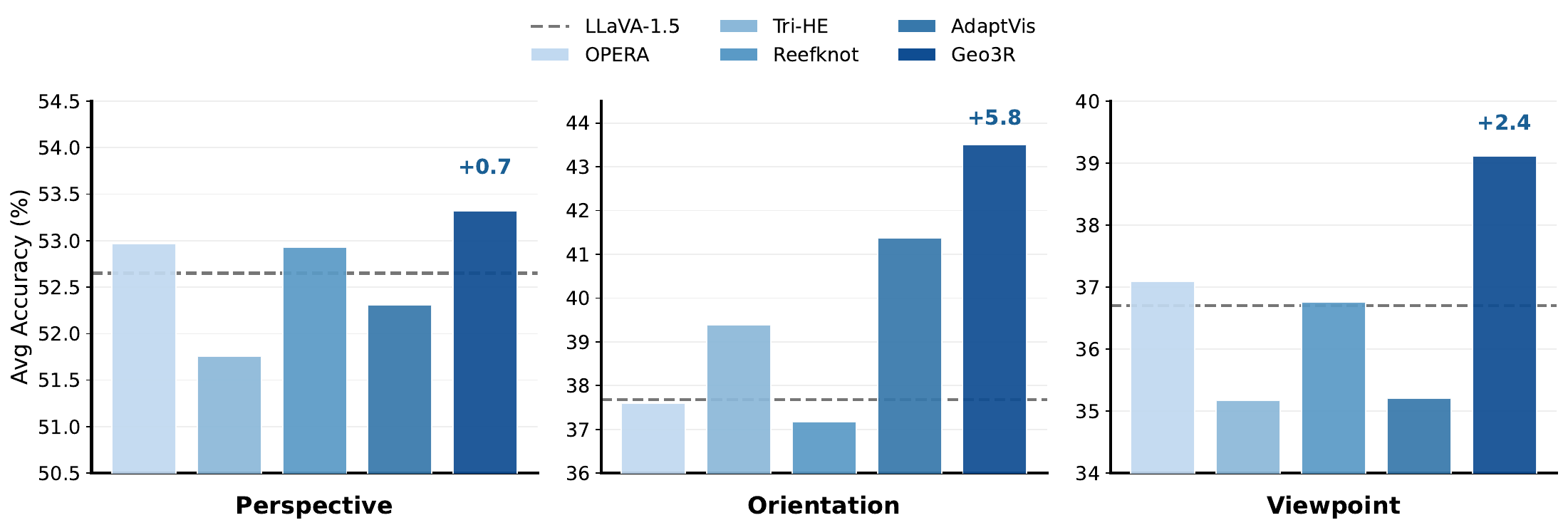}
\caption{Scenario-level average accuracy of hallucination mitigation methods applied to LLaVA-1.5.}
\label{fig:bar_hallucination}
\end{figure}

Table~\ref{tab:ablation_merged} compares existing hallucination mitigation methods and APC-VLM applied to LLaVA-1.5 and Qwen3-VL respectively, and reports the effect of adding Geo3R across three base models.

\textbf{Existing hallucination mitigation methods are ineffective for spatial reasoning.}
In the LLaVA-1.5 group, OPERA, Tri-HE, and Reefknot produce negligible or even negative changes in overall accuracy, with gains or losses within 0.25\%.
AdaptVis achieves a modest 0.69\% gain overall but shows inconsistent per-task behavior, improving person orientation from 51.61\% to 59.24\% while degrading location above from 56.72\% to 46.46\%.
These results suggest that existing methods, which primarily target object-level or relation-level hallucinations through decoding adjustments or attention reallocation, do not address the root cause of spatial reasoning hallucinations, namely the lack of explicit 3D geometric information.
In contrast, Geo3R improves the average accuracy by 2.87\%, achieving the best score on 11 out of 18 tasks.
The largest gain appears on person orientation, which rises from 51.61\% to 67.77\%, where structured geometric evidence provides the MLLM with explicit viewpoint cues that are otherwise unavailable from 2D visual features.
As illustrated in Figure~\ref{fig:bar_hallucination}, this advantage spans all three scenarios, whereas existing methods show marginal or inconsistent improvements.

\textbf{Comparison with tool-augmented reasoning.}
We also compare with APC-VLM~\cite{lee2025perspective}, a concurrent method that renders perspective transformed visual prompts to augment spatial reasoning.
When applied to Qwen3-VL, APC-VLM decreases overall accuracy by 7.92\%, from 55.07\% to 47.15\%.
While it improves on certain viewpoint-related tasks such as orientation on the left and distance, it degrades most others, with location closer to camera dropping by 38.71\% and depth by 20.83\%.
This degradation is largely attributable to its design, which is tailored specifically for perspective-change tasks and relies heavily on small detection models for scene abstraction, causing degraded outputs on non-perspective tasks and when detection fails. In contrast, Geo3R primarily leverages the VLM's own grounding capability and is designed to address all spatial reasoning scenarios, achieving robust improvements across all scenarios.

\textbf{Geo3R scales with base model capability.}
Geo3R yields substantial improvements across all three base models, with gains of 2.87\% on LLaVA-1.5, 10.90\% on Qwen3-VL, and 7.06\% on Gemini-3-Flash.
The lower gain on Gemini-3-Flash compared to Qwen3-VL is partly attributable to its already strong baseline of 65.34\%, which leaves less room for improvement than Qwen3-VL's 55.07\%.
Since Geo3R provides identical geometric evidence regardless of the base model, the difference in gains reflects each model's ability to integrate structured 3D cues with its own visual understanding.
Stronger models such as Qwen3-VL and Gemini-3-Flash can effectively synthesize depth, orientation, and camera parameters with visual context, translating geometric evidence into correct spatial judgments.
In contrast, LLaVA-1.5 is bottlenecked by limited reasoning, making it less able to leverage the additional geometric information.
This scaling behavior suggests that the effectiveness of geometric augmentation is closely tied to the base model's overall capacity, including but not limited to instruction following, structured input integration, and resilience to estimation noise.

\subsection{Ablation Studies}
\label{sec:ablation}

\begin{table}[t]
\caption{Component ablation with Qwen3-VL on 3DSR. Scenario columns report macro-averaged accuracy within each group. Avg is the macro average across all 12 tasks.}
\label{tab:ablation_component}
\centering
\small
\setlength{\tabcolsep}{7pt}
\begin{tabular}{l cccc}
\toprule
Method & Perspective & Orientation & Viewpoint & Avg \\
\midrule
Qwen3-VL & 67.0 & 56.0 & 42.0 & 57.1 \\
+ Geo3R & 77.7 & 66.5 & 54.4 & 68.1 \\
+ Geo3R (a) & 73.2 & 66.1 & 51.0 & 65.3\,{\tiny \textcolor{red}{-2.8}} \\
+ Geo3R (b) & 75.7 & 66.4 & 53.7 & 67.1\,{\tiny \textcolor{red}{-1.0}} \\
+ Geo3R (c) & 69.2 & 55.2 & 59.4 & 62.1\,{\tiny \textcolor{red}{-6.0}} \\
+ Geo3R (d) & 77.4 & 67.1 & 41.7 & 65.0\,{\tiny \textcolor{red}{-3.1}} \\
+ Geo3R (e) & 69.1 & 55.3 & 43.4 & 58.1\,{\tiny \textcolor{red}{-10.0}} \\
\bottomrule
\end{tabular}
\end{table}

\begin{table}[t]
\caption{Effect of removing image input in Stage~2 (w/o Img). Each dataset column reports the macro average across its constituent tasks. Avg is the mean of the three dataset-level values.}
\label{tab:noimg_ablation}
\centering
\small
\begin{tabular}{ll cccc}
\toprule
Model & Method & 3DSR & VSB & CVB & Avg \\
\midrule
Qwen3-VL & Full & 68.1 & 52.2 & 80.5 & 67.0 \\
 & w/o Image & 69.1 & 50.0 & 76.9 & 65.4\,{\tiny \textcolor{red}{-1.6}} \\
\midrule
Gemini-3-Flash & Full & 75.3 & 53.2 & 93.3 & 73.9 \\
 & w/o Image & 75.2 & 50.4 & 91.4 & 72.3\,{\tiny \textcolor{red}{-1.6}} \\
\midrule
LLaVA-1.5 & Full & 48.0 & 36.4 & 53.9 & 46.1 \\
 & w/o Image & 47.6 & 34.2 & 51.9 & 44.6\,{\tiny \textcolor{red}{-1.5}} \\
\bottomrule
\end{tabular}
\end{table}

\textbf{Component ablation.}
Table~\ref{tab:ablation_component} reports ablation results on 3DSR with Qwen3-VL. The full Geo3R pipeline achieves 68.1\% overall accuracy. We evaluate five ablated variants:
(a)~Replacing self-grounding with Grounding DINO.
(b)~Removing SAM segmentation.
(c)~Removing the World Card.
(d)~Removing the Object-Local Card.
(e)~Removing both cards.

Replacing self-grounding with Grounding DINO~(a) reduces accuracy to 65.3\%, with the largest drop on Perspective tasks from 77.7\% to 73.2\%, confirming that the built-in localization of stronger base models is better calibrated with their visual representations. Removing SAM segmentation~(b) causes a smaller degradation to 67.1\%. Perspective and Viewpoint tasks are most affected, dropping from 77.7\% to 75.7\% and from 54.4\% to 53.7\% respectively, because mask-level boundaries directly affect the quality of back-projected point clouds used for depth, extent, and centroid computation. Orientation tasks remain relatively stable at 66.4\%, as orientation estimation relies on cropped image regions rather than precise segmentation masks.
Removing the World Card~(c) causes a 6.0\% overall drop, with Perspective dropping by 8.5\% and Orientation by 11.3\%, as these tasks rely on global spatial context such as camera-relative depth and inter-object distances. Notably, Viewpoint improves slightly, likely because the model no longer needs to select the appropriate card and can focus on the remaining Object-Local Card. Removing the Object-Local Card~(d) leads to a smaller overall drop of 3.1\%, but disproportionately affects Viewpoint tasks by 12.7\%, because viewpoint reasoning depends on object-centric orientation cues. Removing both cards~(e) results in a 10.0\% overall degradation, approaching the unaugmented baseline of 57.1\%, confirming that the two cards provide complementary geometric information and are jointly responsible for the majority of Geo3R's improvements.

\textbf{Effect of image input in Stage~2.}
Table~\ref{tab:noimg_ablation} examines whether the annotated image is necessary in the reasoning stage by removing it and relying solely on structured geometric cards. All three base models show consistent overall degradation of approximately 1.5\% to 1.6\%. Notably, the impact varies across benchmarks. On 3DSR, accuracy remains stable or even slightly improves for Qwen3-VL, from 68.1\% to 69.1\%, suggesting that the image may introduce visual cues that conflict with the geometric cards in Qwen3-VL's reasoning, and removing it eliminates this interference. Across all other model-benchmark combinations, removing the image consistently degrades performance, as visual cues in the annotated image complement geometric measurements for tasks requiring perspective comparison or viewpoint reasoning. Overall, these results confirm that the annotated image remains beneficial and provides complementary information to the geometric cards.

Additional component ablation results on LLaVA-1.5 are provided in the appendix.

\section{Limitations and Conclusion}
\label{sec:conclusion}

We acknowledge several limitations of the current work. First, Geo3R relies on existing geometric estimation tools whose accuracy upper-bounds the framework's performance. However, the modular design allows straightforward replacement as more accurate estimators become available. Second, the framework introduces additional inference-time overhead, but our timing analysis (provided in the appendix) shows it remains comparable to existing methods and can be further reduced with lighter-weight components. Third, the current evaluation focuses on single-image spatial reasoning, though extending to multi-view or video-based scenarios is a natural next step given the framework's geometric foundation.

In summary, we define spatial reasoning hallucination as a subcategory of relation hallucination in MLLMs and propose Geo3R, a training-free, plug-and-play framework that bridges the gap between 2D visual representations and 3D spatial reality through explicit geometric evidence. Extensive experiments show that Geo3R substantially reduces spatial reasoning hallucination across diverse MLLMs, improving Gemini-3-Flash by 7.06\% and Qwen3-VL-8B by 10.90\%. Our results suggest that a key bottleneck for spatial reasoning is the absence of structured 3D geometric information, which can be effectively supplied at inference time without additional training.

\bibliographystyle{ACM-Reference-Format}
\bibliography{sample-base}

\clearpage
\twocolumn[
  \vspace*{1em}
  \begin{center}
  {\LARGE\bfseries Appendix\par}
  \end{center}
  \vspace{1.5em}
]
\appendix
\suppressfloats[t]

\section{Task Descriptions}
\label{sec:appendix_tasks}

Table~\ref{tab:task_descriptions} lists all 18 spatial reasoning tasks evaluated in our experiments, grouped by scenario. Each task is associated with one of three benchmarks: 3DSRBench (3DSR)\cite{ma20253dsrbench}, ViewSpatial-Bench (VSB)\cite{li2025viewspatial}, or CV-Bench (CVB)\cite{tong2024cambriancvbench}. The table also provides the main-paper abbreviation in parentheses and the number of evaluation samples for each task.

\begin{table}[t]
\caption{All 18 spatial reasoning tasks evaluated in this work, grouped by scenario. Abbreviations match those used in the main paper.}
\label{tab:task_descriptions}
\centering
\small
\setlength{\tabcolsep}{3pt}
\begin{tabular}{l p{4.6cm} l r}
\toprule
Scenario & Task & Source & Samples \\
\midrule
\multirow{7}{*}{Perspective}
 & location\_above (Loc. Above) & 3DSR & 1{,}384 \\
 & height\_higher (Hgt. Higher) & 3DSR & 1{,}380 \\
 & location\_closer\_to\_camera (Loc. Cam.) & 3DSR & 1{,}356 \\
 & location\_next\_to (Loc. Next.) & 3DSR & 678 \\
 & multi\_object\_closer\_to (Loc. Obj.) & 3DSR & 700 \\
 & Depth (Depth) & CVB & 600 \\
 & Distance (Dist.) & CVB & 600 \\
\midrule
\multirow{6}{*}{Orientation}
 & cam\_orientation (Cam. Orient.) & VSB & 996 \\
 & person\_orientation (Psn. Orient.) & VSB & 996 \\
 & orientation\_viewpoint (Orient. View.) & 3DSR & 1{,}372 \\
 & multi\_object\_facing (Orient. Facing) & 3DSR & 692 \\
 & \shortstack[l]{multi\_object\_same\_direction\\(Orient. Same-Dir.)} & 3DSR & 688 \\
 & multi\_object\_parallel (Orient. Parallel) & 3DSR & 678 \\
\midrule
\multirow{5}{*}{Viewpoint}
 & \shortstack[l]{multi\_object\_viewpoint\_towards\_object\\(Orient. Twd.)} & 3DSR & 1{,}372 \\
 & orientation\_on\_the\_left (Orient. Left) & 3DSR & 698 \\
 & orientation\_in\_front\_of (Orient. Front) & 3DSR & 688 \\
 & person\_relative\_dir (Psn. Rel-Dir.) & VSB & 842 \\
 & cam\_relative\_dir (Cam. Rel-Dir.) & VSB & 1{,}773 \\
\bottomrule
\end{tabular}
\end{table}

\section{Implementation Details}
\label{sec:appendix_implementation}

\subsection{Base Models}
We evaluate Geo3R on three base MLLMs:
\begin{itemize}
\setlength{\itemsep}{1pt}\setlength{\parsep}{0pt}
    \item \textbf{Qwen3-VL-8B}\cite{yang2025qwen3}: Uses native visual grounding to produce bounding boxes directly, without an external detector.
    \item \textbf{Gemini-3-Flash}\cite{team2023gemini}: Accessed via the official API. Uses native visual grounding for bounding-box extraction.
    \item \textbf{LLaVA-1.5-7B}\cite{llava-1.5}: Uses Grounding DINO\cite{liu2024grounding} for bounding-box extraction and SAM for mask refinement.
\end{itemize}

\subsection{Geometric Estimation Tools}
The geometric evidence extraction stage relies on four pretrained tools:
\begin{itemize}
\setlength{\itemsep}{1pt}\setlength{\parsep}{0pt}
    \item \textbf{DepthPro}~\cite{bochkovskii2024depth}: Produces depth maps from a single image.
    \item \textbf{GeoCalib}~\cite{veicht2024geocalib}: Estimates focal length and camera orientation (roll and pitch) relative to gravity.
    \item \textbf{SAM}~\cite{kirillov2023segment}: Refines bounding boxes into segmentation masks for precise point cloud extraction.
    \item \textbf{Orient Anything V2}~\cite{wang2026orient}: Estimates 3D object orientation from a cropped image region, providing a rotation matrix aligned with world coordinates.
\end{itemize}
All experiments are conducted on NVIDIA H20 and B20Z GPUs.

\subsection{Model-Specific Adaptations}
Table~\ref{tab:model_adaptations} summarizes the key implementation differences across the three base models.

\begin{table}[t]
\caption{Implementation differences across the three base models in Geo3R. Grounding refers to the bounding-box extraction method. Geo card style indicates whether the geometric card uses concise or verbose summaries.}
\label{tab:model_adaptations}
\centering
\small
\setlength{\tabcolsep}{4pt}
\begin{tabular}{l ccc}
\toprule
 & Qwen3-VL & Gemini-3-Flash & LLaVA-1.5 \\
\midrule
Grounding & Native VLM & Native VLM & Grounding DINO \\
SAM masks & \checkmark & \checkmark & \checkmark \\
Stage~1 format & Plain text & JSON & JSON \\
Stage~2 format & Plain text & JSON & JSON \\
Geo card style & Concise & Concise & Verbose \\
\bottomrule
\end{tabular}
\end{table}

For Qwen3-VL and Gemini-3-Flash, each card entry contains a concise \texttt{summary} field alongside the individual numeric fields (e.g., \texttt{extent\_3d}, \texttt{local\_offset}). The summary provides a brief natural-language description such as ``B extends higher in world Y, is farther from the camera, and is offset in top-down view compared with A'', while the individual fields supply precise values for reference.

For LLaVA-1.5, we use a verbose summary that absorbs all individual fields directly into the \texttt{summary} text, rather than listing them separately. The verbose summary additionally includes relational fields such as heading relation, camera-relative direction, and object-local direction. The rationale is that LLaVA-1.5, as a weaker base model, tends to struggle with structured multi-field inputs. Consolidating all information into a single self-contained sentence reduces the risk that the model overlooks key cues and lowers the reasoning burden.

\subsection{Baselines}
We compare against three groups of baselines. All open-source models are evaluated in a zero-shot setting with greedy decoding and default settings unless otherwise noted.

\smallskip
\noindent\textbf{General-purpose MLLMs.}
InternVL3\cite{zhu2025internvl3} uses dynamic resolution (max~12 tiles at 448\,px).
Qwen3-VL disables its thinking mode to ensure direct answer generation.
Kimi-VL\cite{team2025kimi} requires 2{,}048 max output tokens to accommodate its thinking trace, which is stripped before evaluation.

\smallskip
\noindent\textbf{Spatial understanding models.}
SpatialRGPT\cite{cheng2024spatialrgpt} supports optional depth and mask inputs. We set both to \texttt{None} so that it relies solely on RGB, matching all other baselines.
SpatialThinker\cite{batra2025spatialthinker} requires 1{,}024 max output tokens to accommodate its chain-of-thought trace.
All other models use default inference settings.

\smallskip
\noindent\textbf{Hallucination mitigation and tool-augmented methods.}
The first four methods are applied to LLaVA-1.5-7B, the only model supported by all their official implementations.
\begin{itemize}
\setlength{\itemsep}{1pt}\setlength{\parsep}{0pt}
    \item \textbf{OPERA}~\cite{huang2024opera}: Penalizes repetitive attention patterns during beam search. Beam size 5, scale factor 50, threshold 15. \mnt{64}.
    \item \textbf{Tri-HE}~\cite{trihe}: Two-pass self-alignment. Pass~1 (with image) generates a caption. Pass~2 (text-only) answers using only the caption. \mnt{128}.
    \item \textbf{Reefknot}~\cite{zheng2025reefknot}: Dynamic Token Compression on visual tokens. $\alpha{=}0.1$, threshold${=}0.9$, layer${=}38$. \mnt{16}.
    \item \textbf{AdaptVis}~\cite{chen2025spatial}: Adaptive visual attention weighting. Measures output uncertainty and applies weight~1.5 if uncertain ({$>$}0.4), otherwise~0.5. \mnt{64}.
    \item \textbf{APC-VLM}~\cite{lee2025perspective}: Training-free tool-augmented method that builds 3D scene abstraction via GroundingDINO, SAM, DepthPro, and OrientAnything. Evaluated on Qwen3-VL-8B. \mnt{512}.
\end{itemize}

\section{Geometric Card Design}
\label{sec:appendix_card_examples}
\label{sec:appendix_prompts}

\subsection{Card Fields and Pipeline Walkthrough}

Figure~\ref{fig:pipeline_example} illustrates the full Geo3R pipeline on a concrete example drawn from the \texttt{height\_higher} task of 3DSRBench.
All three cards use a \emph{fixed} set of fields (Table~\ref{tab:card_fields}) that are computed for every query regardless of question type. The MLLM selects the relevant evidence at reasoning time.

\definecolor{inputfill}{RGB}{242,242,242}
\definecolor{s1fill}{RGB}{232,240,254}
\definecolor{s1title}{RGB}{210,225,250}
\definecolor{geofill}{RGB}{232,247,232}
\definecolor{geotitle}{RGB}{210,238,210}
\definecolor{s2fill}{RGB}{254,241,230}
\definecolor{s2title}{RGB}{248,225,205}
\definecolor{fieldblue}{RGB}{30,70,130}
\definecolor{fieldgreen}{RGB}{20,90,40}
\newcommand{\fld}[1]{\textrm{\textcolor{fieldblue}{\textbf{#1}}}}
\newcommand{\flg}[1]{\textrm{\textcolor{fieldgreen}{\textbf{#1}}}}

\begin{figure*}[t]
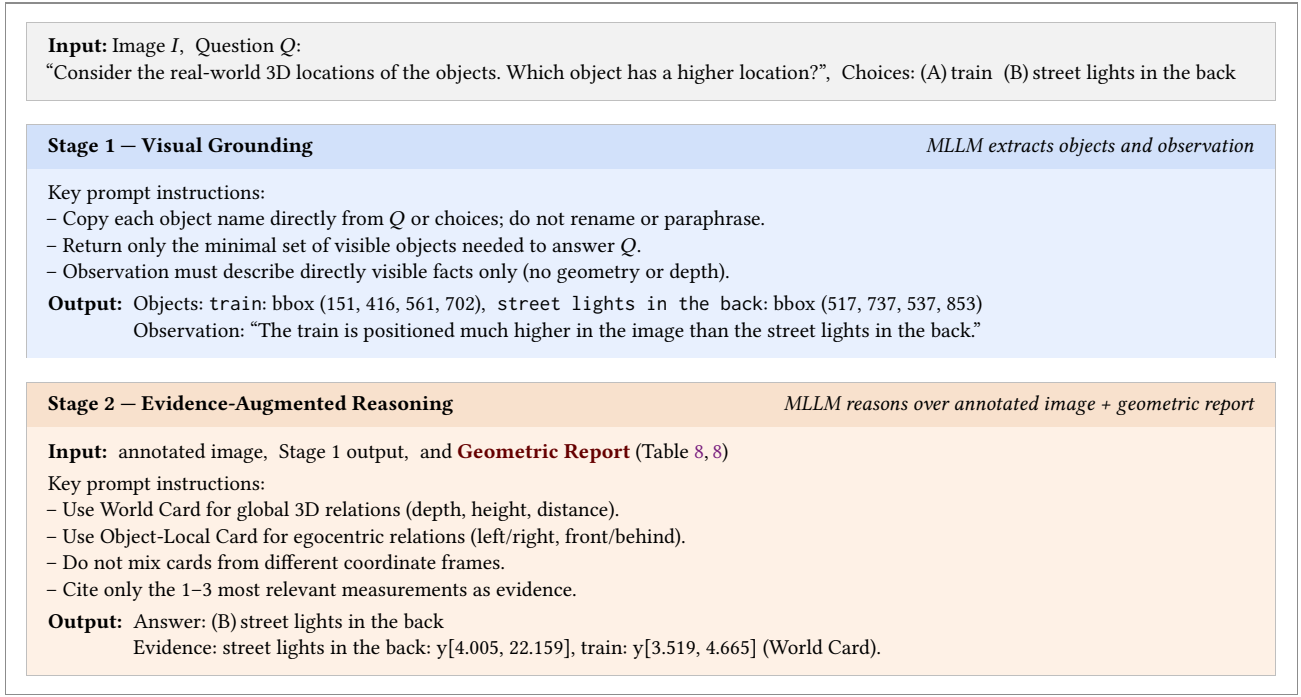

\centering
\begin{tcolorbox}[colback=white, colframe=black!45, boxrule=0.4pt,
  sharp corners, left=5pt, right=5pt, top=4pt, bottom=4pt,
  width=0.96\textwidth]
\small

\begin{tcolorbox}[colback=inputfill, colframe=black!25, boxrule=0.3pt,
  sharp corners, left=5pt, right=5pt, top=3pt, bottom=3pt]
\textbf{Input:} Image $I$,\; Question $Q$: \\``Consider the real-world 3D locations of the objects. Which object has a higher location?'',\; Choices: (A)\,train\; (B)\,street lights in the back
\end{tcolorbox}

\vspace{-1pt}
\begin{tcolorbox}[colback=s1fill, colframe=black!25, boxrule=0.3pt,
  sharp corners, left=5pt, right=5pt, top=3pt, bottom=3pt,
  fonttitle=\small\bfseries, title=Stage~1 --- Visual Grounding,
  toptitle=1.5pt, bottomtitle=1.5pt,
  coltitle=black, colbacktitle=s1title,
  after title={\hfill\normalfont\itshape MLLM extracts objects and observation}]
Key prompt instructions:\\
\quad-- Copy each object name directly from $Q$ or choices; do not rename or paraphrase.\\
\quad-- Return only the minimal set of visible objects needed to answer $Q$.\\
\quad-- Observation must describe directly visible facts only (no geometry or depth).\\[2pt]
\textbf{Output:}\; Objects: \texttt{train}: bbox (151, 416, 561, 702),\; \texttt{street lights in the back}: bbox (517, 737, 537, 853)\\
\phantom{\textbf{Output:}}\; Observation: ``The train is positioned much higher in the image than the street lights in the back.''
\end{tcolorbox}

\vspace{-1pt}
\begin{tcolorbox}[colback=s2fill, colframe=black!25, boxrule=0.3pt,
  sharp corners, left=5pt, right=5pt, top=3pt, bottom=3pt,
  fonttitle=\small\bfseries, title=Stage~2 --- Evidence-Augmented Reasoning,
  toptitle=1.5pt, bottomtitle=1.5pt,
  coltitle=black, colbacktitle=s2title,
  after title={\hfill\normalfont\itshape MLLM reasons over annotated image + geometric report}]
\textbf{Input:}\; annotated image,\; Stage~1 output,\; and \textbf{\textcolor{red!40!black}{Geometric Report}} (Table~\ref{tab:card_fields},\,\ref{tab:card_example})\\[2pt]
Key prompt instructions:\\
\quad-- Use World Card for global 3D relations (depth, height, distance).\\
\quad-- Use Object-Local Card for egocentric relations (left/right, front/behind).\\
\quad-- Do not mix cards from different coordinate frames.\\
\quad-- Cite only the 1--3 most relevant measurements as evidence.\\[2pt]
\textbf{Output:}\; Answer: (B)\,street lights in the back\\
\phantom{\textbf{Output:}}\; Evidence: street lights in the back: y[4.005, 22.159], train: y[3.519, 4.665] (World Card).
\end{tcolorbox}

\end{tcolorbox}
\caption{Geo3R two-stage pipeline on a \texttt{height\_higher} example from 3DSRBench, with key prompt instructions shown inline. All outputs are from an actual Qwen3-VL run.}
\label{fig:pipeline_example}
\end{figure*}

To provide comprehensive geometric evidence, the fields are designed to cover four complementary aspects: 3D position, orientation and visibility, relative direction, and natural-language summary. Table~\ref{tab:card_fields} maps each field to its card and aspect.

\begin{table}[t]
\caption{Fields of the World Card and Object-Local Card, grouped by the spatial aspect they capture. W = World Card, OL = Object-Local Card.}
\label{tab:card_fields}
\centering
\footnotesize
\setlength{\tabcolsep}{2pt}
\begin{tabularx}{\columnwidth}{l l c X}
\toprule
Aspect & Field & Card & Purpose \\
\midrule
\multirow{4}{*}{Position}
 & \texttt{extent\_3d}   & W  & 3D bounding box \\
 & \texttt{raw\_depth}   & W  & Depth to camera \\
 & \texttt{distance\_3d} & W  & Distance between objects \\
 & \texttt{local\_offset} & OL & Offset in ref.\ object's frame \\
\midrule
\multirow{3}{*}{\shortstack[l]{Orient.\\\& vis.}}
 & \texttt{visible\_side\_to\_camera} & W  & Face seen from camera \\
 & \texttt{visible\_side\_to\_target} & OL & Face seen from other obj. \\
 & \texttt{heading\_relation} & W & Same/opposite direction \\
\midrule
\multirow{3}{*}{\shortstack[l]{Relative\\direction}}
 & \texttt{camera\_relative\_direction} & W  & Global-frame direction \\
 & \texttt{object\_local\_direction}    & OL & Egocentric direction \\
 & \texttt{spatial\_relation}  & OL & Vertical (above/below) \\
\midrule
Summary & \texttt{summary} & Both & NL verbalization \\
\bottomrule
\end{tabularx}
\end{table}

We use a full 3D bounding box rather than a single centroid so that the MLLM can compare heights, lateral extents, and depth spans directly.
Orientation fields address tasks where positional information alone is insufficient, such as determining which direction an object faces or which side of one object is visible from another.
Directions are expressed in two frames, global and egocentric, to support both camera-centric and viewpoint-dependent questions.
Each card also includes a natural-language summary that verbalizes key comparisons, reducing the need for the MLLM to interpret raw numbers.

Figure~\ref{tab:card_example} shows the complete geometric report produced for this example. Coordinate convention: $Y_W$ points upward against gravity, so a higher $y_\text{top}$ indicates a physically higher object. Smaller \texttt{raw\_depth} values indicate closer objects.

\begin{figure*}[t]
\centering
\begin{tcolorbox}[colback=white, colframe=black!45, boxrule=0.4pt,
  sharp corners, left=2pt, right=2pt, top=1pt, bottom=1pt,
  width=0.96\textwidth]

\begin{tcolorbox}[colback=inputfill, colframe=black!25, boxrule=0.3pt,
  sharp corners, left=4pt, right=4pt, top=1pt, bottom=1pt,
  fonttitle=\small\bfseries, title=Object Card,
  toptitle=1.5pt, bottomtitle=1.5pt,
  coltitle=black, colbacktitle=black!10]
\small\ttfamily
- train: red box\newline
- street lights in the back: blue box
\end{tcolorbox}

\vspace{-1pt}
\begin{tcolorbox}[colback=s1fill, colframe=black!25, boxrule=0.3pt,
  sharp corners, left=4pt, right=4pt, top=1pt, bottom=1pt,
  fonttitle=\small\bfseries, title=World Card,
  toptitle=1.5pt, bottomtitle=1.5pt,
  coltitle=black, colbacktitle=s1title]
\small\ttfamily
\textrm{\textbf{\textit{Per-object:}}}\newline
- train:\newline
\quad \fld{summary} = spans from left x=-5.437 to right x=-0.041, bottom y=3.519 to top y=4.665, near z=13.324 to far z=18.373 in the world frame, with camera depth 15.509 and side visible from the camera front\newline
\quad \fld{extent\_3d} = x[-5.437, -0.041], y[3.519, 4.665], z[13.324, 18.373]\newline
\quad \fld{raw\_depth} = 15.509;\; \fld{visible\_side\_to\_camera} = front\newline
- street lights in the back:\newline
\quad \fld{summary} = spans from left x=36.417 to right x=49.356, bottom y=4.005 to top y=22.159, near z=99.589 to far z=134.318 in the world frame, with camera depth 116.225 and side visible from the camera front\newline
\quad \fld{extent\_3d} = x[36.417, 49.356], y[4.005, 22.159], z[99.589, 134.318]\newline
\quad \fld{raw\_depth} = 116.225;\; \fld{visible\_side\_to\_camera} = front\par
\vspace{2pt}
\textrm{\textbf{\textit{Pairwise (Reference train $\to$ Target street lights in the back):}}}\newline
\quad \fld{summary} = street lights in the back extends higher in world Y, is farther from the camera, and is offset in top-down view when compared with train\newline
\quad \fld{distance\_3d} = 112.192\newline
\quad \fld{heading\_relation} = same-direction\newline
\quad \fld{camera\_relative\_direction} = front-right\par
\vspace{2pt}\hrule\vspace{2pt}
\textrm{\textbf{\textit{Field Notes:}}}\newline
\fld{extent\_3d}: world frame, camera center as origin, X=right, Y=up, Z=forward. \fld{raw\_depth}: smaller = closer to camera.\newline
\fld{visible\_side\_to\_camera}: side most directly facing the camera. \fld{distance\_3d}: absolute 3D distance.\newline
\fld{camera\_relative\_direction}: target direction from reference in camera frame.\newline
\fld{heading\_relation}: same/opposite = parallel; cross = perpendicular.
\end{tcolorbox}

\vspace{-1pt}
\begin{tcolorbox}[colback=geofill, colframe=black!25, boxrule=0.3pt,
  sharp corners, left=4pt, right=4pt, top=1pt, bottom=1pt,
  fonttitle=\small\bfseries, title=Object-Local Card,
  toptitle=1.5pt, bottomtitle=1.5pt,
  coltitle=black, colbacktitle=geotitle]
\small\ttfamily
\textrm{\textbf{\textit{Target street lights in the back in Reference train's local frame:}}}\newline
\quad \flg{summary} = In train's local frame, street lights in the back is to the left, behind, and lower than train\newline
\quad \flg{local\_offset} = po\_x=95.614, po\_y=45.557, po\_z=-37.007\newline
\quad \flg{object\_local\_direction} = back-left;\; \flg{visible\_side\_to\_target} = left;\; \flg{spatial\_relation} = below but not aligned\par
\vspace{2pt}
\textrm{\textbf{\textit{Target train in Reference street lights in the back's local frame:}}}\newline
\quad \flg{summary} = In street lights in the back's local frame, train is to the right, in front, and higher than street lights in the back\newline
\quad \flg{local\_offset} = po\_x=-51.346, po\_y=-99.200, po\_z=10.485\newline
\quad \flg{object\_local\_direction} = front-right;\; \flg{visible\_side\_to\_target} = front;\; \flg{spatial\_relation} = above but not aligned\par
\vspace{2pt}\hrule\vspace{2pt}
\textrm{\textbf{\textit{Field Notes:}}}\newline
\flg{local\_offset}: reference object's own frame; po\_x = left/right, po\_y = front/back, po\_z = up/down.\newline
\flg{object\_local\_direction}: overall local direction; \flg{spatial\_relation}: vertical relation + top-down alignment.\newline
\flg{visible\_side\_to\_target}: side of ref.\ object facing target; front/back if |po\_y| >= |po\_x|, else left/right.\newline
Do not treat \flg{local\_offset} as absolute world position, height, or camera depth.
\end{tcolorbox}

\end{tcolorbox}
\caption{Geometric report produced for the example in Figure~\ref{fig:pipeline_example}. For brevity, only one pairwise direction is shown in the World Card; the full report includes both directions.}
\label{tab:card_example}
\end{figure*}

\subsection{Card Selection Strategy}
\label{sec:appendix_card_selection}
All three cards are always included in the geometric report. Rather than filtering cards by task type, we provide usage guidelines within the Stage~2 prompt and let the MLLM select the relevant card based on the question. This avoids the need for a task-type classifier and allows the MLLM to flexibly combine evidence when a question spans multiple spatial reasoning aspects.

\subsection{Eight-Way Direction Quantization}
Azimuth angles are quantized into eight directions by dividing the $360^\circ$ range into $45^\circ$ bins centered at $0^\circ$:

\smallskip
\begin{center}
\small
\begin{tabular}{ll@{\quad}ll}
front & $[337.5^\circ, 22.5^\circ)$ & back & $[157.5^\circ, 202.5^\circ)$ \\
front-left & $[22.5^\circ, 67.5^\circ)$ & back-right & $[202.5^\circ, 247.5^\circ)$ \\
left & $[67.5^\circ, 112.5^\circ)$ & right & $[247.5^\circ, 292.5^\circ)$ \\
back-left & $[112.5^\circ, 157.5^\circ)$ & front-right & $[292.5^\circ, 337.5^\circ)$ \\
\end{tabular}
\end{center}

\noindent This quantization is applied to both camera-relative directions (computed from the displacement vector between object centroids in camera space) and object-local directions (computed from the target position in the reference object's local frame).

\subsection{Object Name Resolution}
\label{sec:appendix_name_resolution_inline}

As noted in Section~4.1, the visual grounding stage must extract object names that faithfully preserve disambiguating context from the question (e.g., ``the lamppost in the back'' rather than simply ``lamppost''). We achieve this through a combination of prompt design and post-processing.

\textbf{Prompt design.}
The Stage~1 prompt explicitly instructs the MLLM to ``copy each object name directly from the question text or answer choices. Do not rename, merge, or paraphrase.'' This encourages the model to retain distinguishing modifiers such as spatial descriptors, ordinals, or visual attributes that are necessary for unambiguous grounding.

\textbf{Post-processing.}
After the MLLM returns the parsed object list, a reconciliation step validates each name against the source text (question and answer choices) via token-level matching. Names that cannot be matched to any phrase in the source text are discarded, preventing hallucinated or fabricated object references. When multiple extracted names are subsumed by a longer phrase (e.g., ``car'' and ``red car''), the more specific name is retained. If all extracted names fail validation, the system falls back to using the answer choice texts directly as object names.

\section{Additional Experiments}
\label{sec:appendix_additional}

\subsection{Grounding Ablation (LLaVA-1.5)}
\label{sec:appendix_ablation}

Table~\ref{tab:ablation_llava} presents grounding ablation results for LLaVA-1.5 across all three benchmarks. Since LLaVA-1.5 lacks native visual grounding, the default configuration uses Grounding DINO with SAM. We compare three grounding configurations:

\begin{table}[t]
\caption{Effect of different bounding-box extraction configurations on LLaVA-1.5 across all three benchmarks. GD = Grounding DINO.}
\label{tab:ablation_llava}
\centering
\small
\setlength{\tabcolsep}{8pt}
\begin{tabular}{l cccc}
\toprule
Variant & 3DSR & VSB & CVB & Avg \\
\midrule
LLaVA-1.5 & 45.9 & 31.8 & 49.9 & 42.6 \\
+ Geo3R, GD+SAM & 48.0 & 36.4 & 53.9 & 46.1\,{\tiny \textcolor{green!60!black}{+3.6}} \\
+ Geo3R, GD w/o SAM & 48.2 & 34.2 & 55.3 & 45.9\,{\tiny \textcolor{green!60!black}{+3.3}} \\
+ Geo3R, VLM+SAM & 48.1 & 33.9 & 54.8 & 45.6\,{\tiny \textcolor{green!60!black}{+3.0}} \\
\bottomrule
\end{tabular}
\end{table}

The full pipeline with Grounding DINO and SAM (GD+SAM) achieves the best overall accuracy of 46.1\%.
Removing SAM (GD w/o SAM) slightly reduces accuracy to 45.9\%, with VSB showing the largest degradation from 36.4\% to 34.2\%, while 3DSR and CVB remain stable.
Using LLaVA-1.5's native grounding with SAM (VLM+SAM) yields 45.6\%, lower than with GD+SAM, which reflects that in the case of LLaVA-1.5, native grounding produces less accurate bounding boxes than Grounding DINO.
All three configurations outperform the unaugmented LLaVA-1.5 baseline of 42.6\%.

\subsection{Inference Efficiency}
\label{sec:appendix_efficiency}

\begin{table}[t]
\caption{Per-sample inference time (mean\,$\pm$\,std, in seconds) measured on a single NVIDIA H20 GPU. Each method is evaluated on 60 samples per task across all 12 3DSRBench tasks, for a total of 720 samples. GD = Grounding DINO, VLM = native grounding.}
\label{tab:efficiency}
\centering
\small
\setlength{\tabcolsep}{5pt}
\begin{tabular}{l cccc}
\toprule
 & +AdaptVis & +APC-VLM & +Geo3R (VLM) & +Geo3R (GD) \\
\midrule
LLaVA-1.5 & 3.51{\tiny$\pm$1.38} & --                   & 5.46{\tiny$\pm$1.76} & 4.23{\tiny$\pm$0.56} \\
Qwen3-VL  & --                   & 7.93{\tiny$\pm$6.28} & 8.15{\tiny$\pm$1.90} & 5.76{\tiny$\pm$1.39} \\
\bottomrule
\end{tabular}
\end{table}

Table~\ref{tab:efficiency} reports the per-sample inference time of Geo3R and two comparison methods: AdaptVis, a hallucination mitigation baseline evaluated on LLaVA-1.5, and APC-VLM, a tool-augmented spatial reasoning method evaluated on Qwen3-VL.

On LLaVA-1.5, Geo3R with Grounding DINO achieves comparable latency to AdaptVis at 4.23s versus 3.51s, while delivering substantially larger accuracy improvements. Using VLM-based grounding with SAM moderately increases the cost to 5.46s but avoids the external detector entirely.

On Qwen3-VL, Geo3R (GD) achieves 5.76s with a standard deviation of $\pm$1.39 and an overall accuracy of 65.3\% on 3DSR, representing only a 2.8\% drop from the VLM grounding variant (68.1\% at 8.15s). Both configurations exhibit stable and predictable runtimes, because Geo3R invokes the base VLM exactly twice, once in Stage~1 and once in Stage~2. APC-VLM, by contrast, requires 7.93s with a standard deviation of $\pm$6.28, as each of its three pipeline stages involves multiple VLM calls whose cost grows with the number of detected objects.

As a training-free framework, Geo3R achieves accuracy substantially above spatially fine-tuned models such as SpatialThinker and SenseNova-SI~\cite{sensenova-si}, while its inference time is comparable to APC-VLM and can be reduced further by substituting Grounding DINO for VLM-based grounding, which also eliminates the VLM call in Stage~1 and reduces the pipeline to a single VLM inference in Stage~2. Furthermore, both Geo3R variants remain far below dense reconstruction-based perspective reasoning approaches, which according to APC-VLM's own evaluation require over ten times its inference time. Finally, since each geometric estimation tool is an independent module, it can be swapped for a faster or more accurate alternative without retraining, providing a clear path to further latency reduction.

\section{Robustness Analysis}
\label{sec:appendix_robustness}

To analyze the robustness of Geo3R, we conduct three levels of experiments. First, we measure tool success rates across LLaVA-1.5, Qwen3-VL, and Gemini-3-Flash, covering 17{,}493 samples per model. Second, we manually annotate 100 images from 3DSRBench on Qwen3-VL (196 task instances across all 12 tasks) to verify tool correctness beyond success rate. Third, we inject synthetic noise into Geometric Cards on 3DSRBench with Qwen3-VL to test robustness.

\begin{table}[t]
\caption{Tool success rate and accuracy. $^{*}$After excluding upstream grounding/SAM errors. $^{\ddag}$Grounding success rate includes VLM detection with feedback retry.}
\label{tab:tool_acc}
\centering
\small
\setlength{\tabcolsep}{4pt}
\begin{tabular}{l c c c}
\toprule
Tool & Succ.\ Rate (\%) & Accuracy (\%) & Own Acc.$^{*}$ (\%) \\
\midrule
Grounding$^{\ddag}$ & 100 & 96.4 & 96.4 \\
SAM Segmentation & 100 & 94.4 & 97.9 \\
Depth Estimation & 100 & 89.2 & 94.8 \\
Orientation Est. & 100 & 71.5 & 75.6 \\
\bottomrule
\end{tabular}
\end{table}

Table~\ref{tab:tool_acc} reports results from the first two levels (GeoCalib is excluded as its outputs are difficult to verify without ground-truth calibration). In terms of success rate, all core tools achieve 100\%, confirming the pipeline reliably produces Geometric Cards.

In terms of accuracy, the only possible cascading path in Geo3R is Grounding $\to$ SAM $\to$ Depth / Orientation, where Depth and Orientation are parallel leaf modules that do not affect each other. Grounding and SAM, as the only upstream stages that can trigger cascading, achieve 96.4\% and 97.9\% own accuracy respectively. Even considering cascading, their combined accuracy remains above 94\%, with minimal impact on downstream modules.

For the downstream leaf modules, Depth achieves 89.2\% end-to-end accuracy and 94.8\% own accuracy, indicating it remains robust even under upstream cascading. Orientation achieves 71.5\% end-to-end and 75.6\% own accuracy, where the gap to other tools is primarily due to the inherent difficulty of monocular orientation estimation rather than cascading from upstream. Nevertheless, orientation still brings consistent positive improvement across all evaluated models, as shown in Section~\ref{sec:appendix_additional_baselines}.

\begin{table}[t]
\caption{Accuracy under noise injection on 3DSRBench (Qwen3-VL). D.$\sigma$ adds Gaussian noise to depth values; D.Swap swaps depth values between two objects; O.Flip reverses the object facing direction in the card.}
\label{tab:noise}
\centering
\small
\setlength{\tabcolsep}{3pt}
\begin{tabular}{l c c c c c c}
\toprule
 & Qwen3 & Geo3R & D.$\sigma$=10\% & D.$\sigma$=50\% & D.Swap & O.Flip \\
\midrule
Avg & 57.1 & 68.1 & 68.1 & 67.9 & 58.8 & 63.9 \\
\bottomrule
\end{tabular}
\end{table}

Table~\ref{tab:noise} reports results from the third level. Gaussian noise up to 50\% causes negligible degradation, confirming the model relies on relative ordering rather than absolute values. Even under more extreme semantic corruption (D.Swap, O.Flip) that damages specific categories, the overall Avg still remains above the no-card baseline, indicating the model retains self-correction ability from its own visual features. Moreover, these corrupted scenarios are synthetically injected and rarely occur in practice, as confirmed by the tool success rates and accuracy reported in Table~\ref{tab:tool_acc}.

\section{Card Format Ablation}
\label{sec:appendix_card_format}

We ablate the Geometric Card design on 3DSRBench with Qwen3-VL to study the effect of each textual component. We evaluate five variants: (a)~w/o usage preamble, (b)~w/o field notes (coordinate explanations), (c)~w/o natural-language summaries, (d)~all fields in JSON format, and (e)~all content as unstructured text.

\begin{table}[t]
\caption{Card format ablation on 3DSRBench with Qwen3-VL.}
\label{tab:card_format}
\centering
\small
\setlength{\tabcolsep}{4pt}
\begin{tabular}{l c c c c c c}
\toprule
 & Full & (a) & (b) & (c) & (d) & (e) \\
\midrule
Avg & 68.1 & 66.5 & 66.0 & 66.1 & 66.2 & 61.2 \\
\bottomrule
\end{tabular}
\end{table}

As shown in Table~\ref{tab:card_format}, removing individual components (a--c) each causes a 1.6--2.1\% drop. Variant (d) retains explicit field names and only drops 1.9\%, while (e) buries all values in sentences without field identifiers, leading to a 6.9\% drop. This confirms that structured format with clear field names helps the MLLM effectively utilize geometric evidence.

\section{Additional Baselines}
\label{sec:appendix_additional_baselines}

We further add the latest general-purpose MLLMs (Qwen3.5-9B, Gemini-3-Pro, GPT-5.5) and spatial understanding models (VST-7B-RL~\cite{yang2025visual}, SpaceR~\cite{ouyang2025spacer}) to provide a more comprehensive comparison, and also evaluate Geo3R on these new general-purpose MLLMs. Table~\ref{tab:baselines} presents representative baselines from our original submission together with newly added models on 3DSRBench.

\begin{table}[t]
\caption{Additional baselines on 3DSRBench. Avg is the macro average over all 12 tasks. $^{\dagger}$Newly added models.}
\label{tab:baselines}
\centering
\small
\setlength{\tabcolsep}{4pt}
\begin{tabular}{l c c c c}
\toprule
Model & Persp. & Orient. & View. & Avg \\
\midrule
\multicolumn{5}{l}{\textit{General-Purpose MLLMs}} \\
Qwen3-VL & 67.0 & 56.0 & 42.0 & 57.1 \\
Gemini-3-Flash & 69.3 & 66.4 & 62.5 & 66.6 \\
GPT-5 & 76.9 & 63.3 & 59.4 & 68.0 \\
Qwen3.5-9B$^{\dagger}$ & 71.7 & 60.8 & 53.6 & 63.5 \\
Gemini-3-Pro$^{\dagger}$ & 77.1 & 66.6 & 65.7 & 70.8 \\
GPT-5.5$^{\dagger}$ & 74.7 & 71.6 & 72.8 & 73.2 \\
\midrule
\multicolumn{5}{l}{\textit{Spatial Understanding Models}} \\
SpatialThinker & 65.5 & 54.3 & 40.1 & 55.4 \\
SpaceR$^{\dagger}$ & 60.4 & 51.9 & 40.9 & 52.7 \\
VST-7B-RL$^{\dagger}$ & 70.5 & 54.7 & 45.0 & 58.9 \\
\midrule
\multicolumn{5}{l}{\textit{Ours}} \\
+ Qwen3-VL & 77.7 & 66.5 & 54.4 & 68.1 \\
+ Gemini-3-Flash & 79.1 & 71.2 & 74.5 & 75.3 \\
+ Qwen3.5-9B$^{\dagger}$ & 79.4 & 67.5 & 73.7 & 74.0 \\
+ Gemini-3-Pro$^{\dagger}$ & 80.5 & 70.9 & 75.5 & 76.0 \\
+ GPT-5.5$^{\dagger}$ & 80.8 & 72.9 & 76.2 & 77.0 \\
\bottomrule
\end{tabular}
\end{table}

Among the newly added spatial understanding models, VST-7B-RL~\cite{yang2025visual} performs best at 58.9\%, surpassing SpatialThinker, the strongest spatial understanding model in our original submission, and also slightly above the Qwen3-VL baseline, while SpaceR~\cite{ouyang2025spacer} achieves only 52.7\%. All spatial understanding models still remain well below Geo3R with Qwen3-VL.

For general-purpose MLLMs, all newly added models show overall improvements over their predecessors. We further evaluate Geo3R on these models. Geo3R with Qwen3.5-9B reaches 74.0\%, improving by 10.5\%, and surpasses both Gemini-3-Pro and GPT-5.5. Even for the strongest models where accuracy is already high, Geo3R still brings consistent improvements, with Gemini-3-Pro rising from 70.8\% to 76.0\% and GPT-5.5 from 73.2\% to 77.0\%, demonstrating that our method remains effective on the latest and strongest models.

We additionally evaluate Geo3R with Qwen3-VL on SpatialScore~\cite{wu2026spatialscore}, a comprehensive spatial reasoning benchmark that aggregates data from OmniSpatial, SPAR-Bench, and 20+ other spatial benchmarks. On its single-image spatial subsets (View, Depth Estimation, 3D Perception), the average accuracy improves from 43.7\% to 50.8\%, further confirming the generality of our approach beyond the three benchmarks in the main paper.

\end{document}